%% file: New_IEEEtran_how-to.tex
\documentclass[lettersize,journal]{IEEEtran}
\usepackage{amsmath,amsfonts}
\usepackage{array}
\usepackage[caption=false,font=normalsize,labelfont=sf,textfont=sf]{subfig}
\usepackage{textcomp}
\usepackage{stfloats}
\usepackage{url}
\usepackage{verbatim}
\usepackage{graphicx}
\usepackage{cite}
\usepackage[colorlinks,
            linkcolor=black,
            anchorcolor=black,
            citecolor=black]{hyperref}
\usepackage{amsmath,amssymb,amsfonts}
\usepackage{graphicx}
\usepackage{textcomp}
\usepackage{xcolor}
\usepackage{stfloats}
\usepackage{array}
\usepackage{booktabs}
\usepackage{svg}
\usepackage{pifont}       
\usepackage{bbding}       
\usepackage{fontawesome}  

\makeatletter
\newif\if@restonecol
\makeatother

\usepackage[linesnumbered,ruled,vlined]{algorithm2e}
\usepackage{algpseudocode}
\usepackage{amsmath}
\usepackage{bm}

\def\BibTeX{{\rm B\kern-.05em{\sc i\kern-.025em b}\kern-.08em
    T\kern-.1667em\lower.7ex\hbox{E}\kern-.125emX}}
\usepackage{balance}
\DeclareRobustCommand*{\IEEEauthorrefmark}[1]{%
    \raisebox{0pt}[0pt][0pt]{\textsuperscript{\footnotesize\ensuremath{#1}}}}
\begin{document}
\title{FTFDNet: Learning to Detect Talking Face Video Manipulation with Tri-Modality Interaction}
\author{\IEEEauthorblockN{
Ganglai Wang\IEEEauthorrefmark{1}\textsuperscript{*},
Peng Zhang\IEEEauthorrefmark{1}\textsuperscript{*}\textsuperscript{$\dagger$},
Junwen Xiong\IEEEauthorrefmark{1}\textsuperscript{*},
Feihan Yang\IEEEauthorrefmark{1},
Wei Huang\IEEEauthorrefmark{2}, and
Yufei Zha\IEEEauthorrefmark{1}}
%

\IEEEauthorblockA{\IEEEauthorrefmark{1}ASGO, School of Computer Science, Northwestern Polytechnical University, Xi'an, China}

\IEEEauthorblockA{\IEEEauthorrefmark{2}China Mobile-NCU AI\&IOT Jointed Lab, Informatization Office, Nanchang University, Nanchang, China} 
\thanks{*Equal contribution as first author.}
\thanks{$\dagger$Corresponding author: Peng Zhang (email: zh0036ng@nwpu.edu.cn).}
}

\maketitle

\begin{abstract}
DeepFake based digital facial forgery is threatening public media security, especially when lip manipulation has been used in talking face generation, and the difficulty of fake video detection is further improved. By only changing lip shape to match the given speech, the facial features of identity are hard to be discriminated in such fake talking face videos. Together with the lack of attention on audio stream as the prior knowledge, the detection failure of fake talking face videos also becomes inevitable. It's found that the optical flow of the fake talking face video is disordered especially in the lip region while the optical flow of the real video changes regularly, which means the motion feature from optical flow is useful to capture manipulation cues. In this study, a fake talking face detection network (FTFDNet) is proposed by incorporating visual, audio and motion features using an efficient cross-modal fusion (CMF) module. Furthermore, a novel audio-visual attention mechanism (AVAM) is proposed to discover more informative features, which can be seamlessly integrated into any audio-visual CNN architecture by modularization. With the additional AVAM, the proposed FTFDNet is able to achieve a better detection performance than other state-of-the-art DeepFake video detection methods not only on the established fake talking face detection dataset (FTFDD) but also on the DeepFake video detection datasets (DFDC and DF-TIMIT).
\end{abstract}


\input{sec_introduction}

\input{sec_ralated_work}

\input{sec_proposed_method}

\input{sec_experiment}

\input{sec_conclusion}
\section*{Acknowledgments}
This work was supported in part by the National Natural Science Foundation of China under Grants 61971352 and 61862043, in part by the Natural Science Foundation of Jiangxi Province under Grant 20204BCJ22011.

\bibliographystyle{IEEEtran}
\bibliography{inference}

\end{document}

%% file: sec_introduction.tex
\section{Introduction}
Human facial features are unique to everyone, as the symbols of personal identity, they play an important role in social communication. Over the last decades, human face forgery with deep neuron networks has been extensively studied \cite{02, 03, 04, 06, 07}, and those manipulation methods are uniformly called DeepFake. By regarding the level of manipulation, DeepFake can be usually categorized into four groups \cite{01}: entire face synthesis, identity swap, attribute manipulation and expression swap. Not only limited for the purpose of data augmentation, these unrealistic face images may wantonly spread on the Internet and cause a series of moral problems. Fortunately, many forgery detection algorithms \cite{08, 09, 10,11,12,13,14,15,16} have been proposed to combat DeepFake, especially for digital facial forgery.

In the age of AI, do you still believe that the voice from a person in the video is realistic or not?  More recently, an emergence of talking face generation \cite{17, 18, 24, 19, 26, 27, 40, 28, 8995571} has posed a new challenge to fake detection. Compared with common DeepFake, talking face generation only manipulates the lip shape to match the given speech, and does not change facial features of identity, which has stronger concealment. As more and more works have achieved accurate lip synchronization, the generation of undistinguishable videos of fake talking faces is no longer difficult. By using these methods, the lip shapes of characters in the video can be easily manipulated, which may help the generated fake talking face videos (e.g. fake news) to spread disinformation and conduct online fraud. Imagine that if there is an email with such a video of assignments from leaders or help from family members, do you believe or not? What you have seen is hard to determine, and `Seeing is Believing' has become a serious challenge.

\begin{figure}[t]
\centering
\centerline{\includegraphics [width=0.48\textwidth]{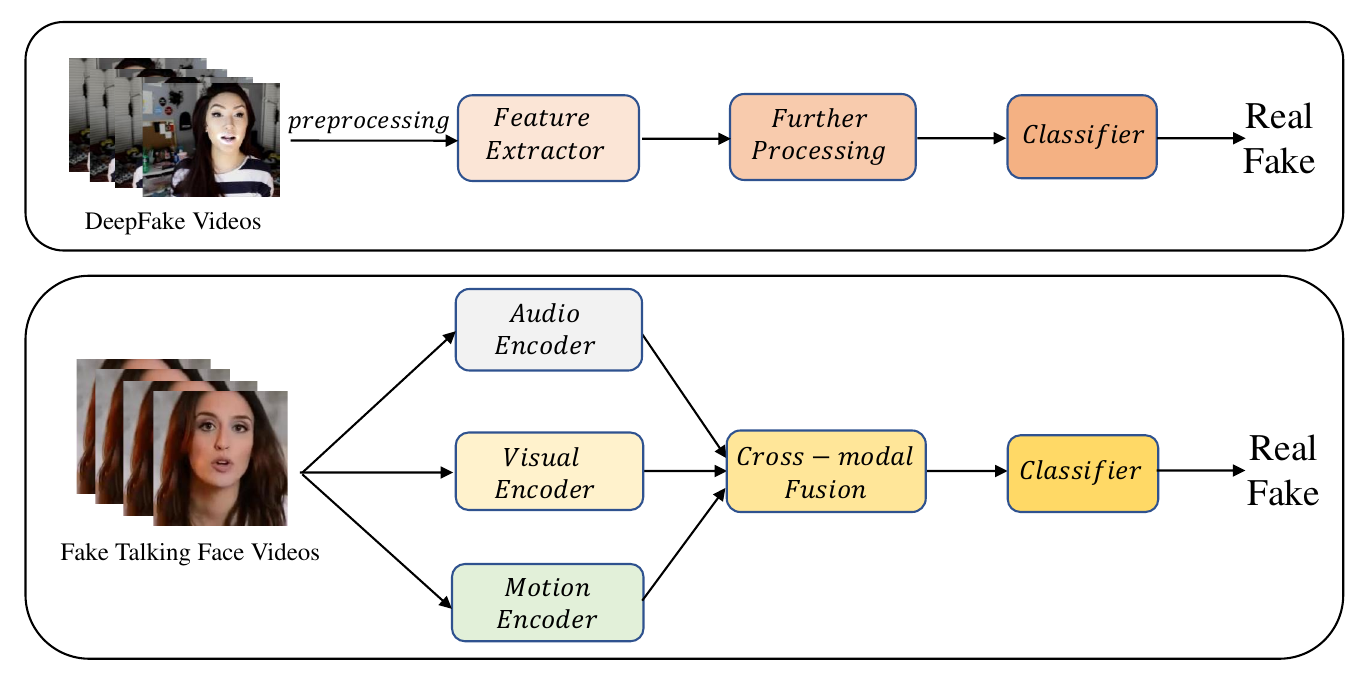}}
\setlength{\abovecaptionskip}{0.cm}
\caption{The process of video forgery detection. Top: the traditional DeepFake video detection process, which only uses the visual modality. Bottom: our multi-modal detection framework, which learns a joint representation from visual, audio and motion features.}
\setlength{\abovecaptionskip}{0.cm}
\label{img_begin}
\end{figure}

\begin{figure*}[ht]
\centering
\centerline{\includegraphics [width=1.05\textwidth]{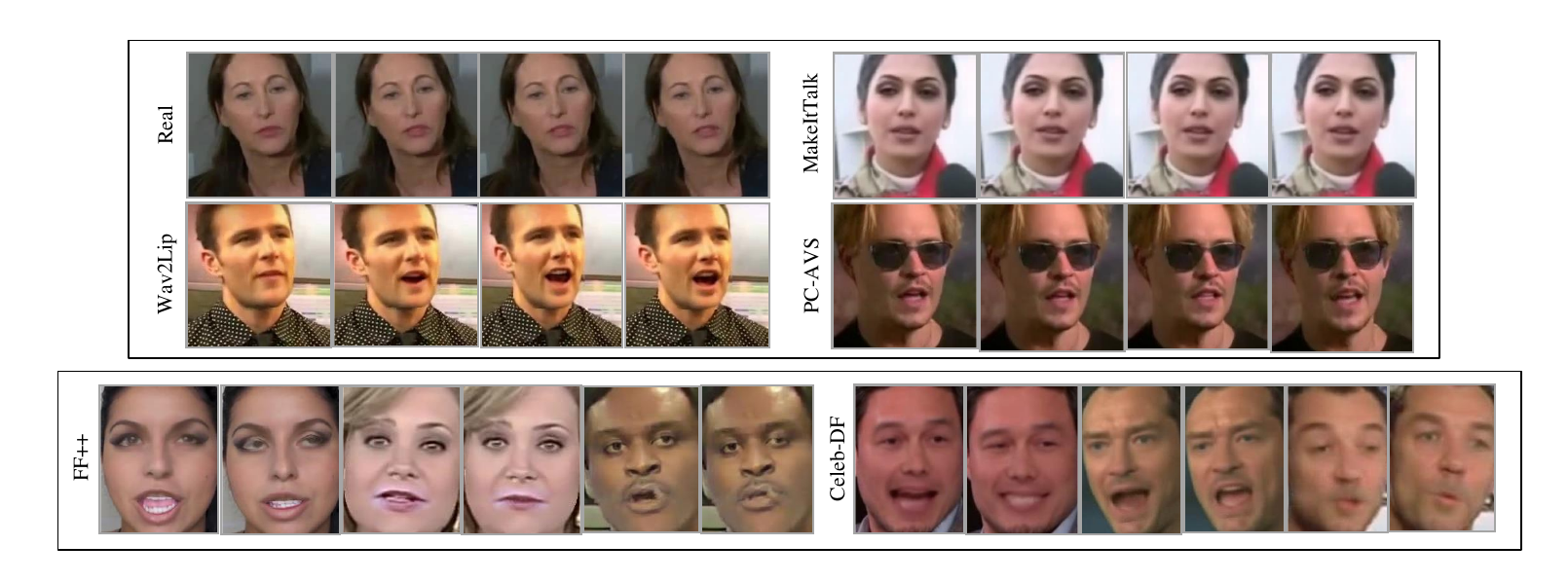}}
\setlength{\abovecaptionskip}{0.cm}
\caption{Examples of face images from FTFDD and DeepFake video detection datasets including FF++ and Celeb-DF. The images on the top are from our established dataset including the original video frames without any manipulation and the fake talking face video frames that are forged by Wav2Lip \cite{26}, MakeItTalk \cite{40} and PC-AVS \cite{28}. The images on the bottom are from available popular DeepFake video detection datasets including FaceForensics++ (FF++)\cite{Rssler2019FaceForensicsLT} and Celeb-DF\cite{Li2019CelebDFAL}.}
\label{img_visual}
\end{figure*}

The majority of existing DeepFake video detection schemes are visual-only based, which typically contain data pre-processing, feature extractor, further processing for effectively utilizing the extracted features and a classifier for obtaining the classification probability, as shown in Fig. \ref{img_begin}(top). Due to the degradation of frame-wise information in video encoding, it is hard for many fake image detection approaches \cite{Farid2008ASO, 9878273, 9878360} to be directly applied to the task of DeepFake video detection, which usually requires to utilize the inter-frame temporal information \cite{20, 21, 22, 23, 9643421}. Unfortunately, the functionality of the audio stream has not been sufficiently exploited in most of current DeepFake video detection studies even it takes an essential part of the video. To capture forgery cues, only few audio-visual based detection methods \cite{Chugh2020NotMF, Chung2016OutOT, Zhou2021JointAD} have attempted to employ the inconsistency between audio and visual modalities. The reasons behind the limited study of audio-visual fusion strategy in traditional DeepFake video detection can be explained as: (i) the shortage of high-quality datasets which contain both audio and visual information; (ii) no direct correlation between the face forgery and video speech in conventional DeepFake, which leads negligible enhancement via audio stream integration. In fact, the operation of talking face forgery is performed by reshaping the character's mouth appearance to match the given speech, which means that the capability of fake video discrimination can be substantially improved with an effective audio guidance.


According to biological perception mechanism, human multisensory neurons in the superior colliculus are capable of combining multi-modal sensory information about the common source, which is able to improve the ability of object localization and discrimination, even to accelerate the response to them\cite{36,37,38,39,40}. When the visual and audio information is received at the same time, the auditory and visual perception systems are activated to send signals to superior colliculus multisensory neurons, in which the separately obtained information is synthesized as the response to the external stimulation. With the combination of the audio-visual features in a class of neurons, this perceptual mechanism enables the informed decisions output based on a fusion-decoding process. It has been also verified that the auditory information can enhance post-sensory visual evidence \cite{41} during rapid multisensory decision-making, which also motivates a study of fake information perception using joint audio-visual representation.

For more challenging talking face video forgery detection with few artifacts of intra-frame with audio-visual hints, the inter-frame motion such as optical flow can be taken into consideration to capture the subtle manipulation artifacts of talking face video manipulation. Fig. \ref{img_flow} is the visualization of the estimated optical flow using Gunnar Farneback' algorithm \cite{Farnebck2003TwoFrameME} on the real and fake video frames. It can be found that the optical flow of the real videos is generally smooth and coherent, in comparison to the frequent disorders in that of the fake talking face videos, especially in the manipulated mouth region area. Considering such an advantage, optical flow based motion feature is also incorporated into the proposed detection model to detect imperceptible differences between real and fake in talking face videos.

%

%



Inspired by the decision-making mechanism of human multisensory perception system, in this work, the \textbf{F}ake \textbf{T}alking \textbf{F}ace \textbf{D}etection \textbf{N}etwork (FTFDNet) is proposed by incorporating visual modality, audio information and motion feature as shown in Fig. \ref{img_begin}(bottom). A self-attention based \textbf{C}ross-\textbf{M}odal \textbf{F}usion (CMF) module is designed to explore the inter-relationships across different modalities, and a novel \textbf{A}udio-\textbf{V}isual \textbf{A}ttention \textbf{M}echanism (AVAM) is also proposed to discover which portion of the feature is more informative to the network. Specifically, the proposed AVAM is end-to-end trainable along with CNNs which is able to be seamlessly embedded into more audio-visual CNN architectures, as functioning in the proposed base network for fake talking face detection. Extensive experiments on our established \textbf{F}ake \textbf{T}alking \textbf{F}ace \textbf{D}etection \textbf{D}ataset (FTFDD), DeepFake video detection datasets (DFDC\cite{Dolhansky2019TheDD} and DFTIMIT\cite{Korshunov2018DeepFakesAN}) have validated a superior performance compared to the other state-of-the-art works. The main contributions are summarized as follows:


\begin{itemize}
\item To fully utilize the inter-relationships between different modalities for forgery detection, a cross-modal fusion scheme is proposed by learning a joint feature representation from audio, visual and motion information.


\item A novel audio-visual attention mechanism is proposed to discover more informative features, which significantly outperforms the popular visual attention mechanism.

\item A large-scale and challenging fake talking face video detection dataset is established, which is generated with state-of-the-art talking face generation methods.

\item In comparison to DeepFake video detection, a more challenging task of fake talking face video detection is introduced with a multi-modal based modeling solution. Extensive experiments demonstrate that the proposed model achieves state-of-the-art performance on different datasets.

\end{itemize}

The rest of the paper is organized as follows: Sec. 2 discusses more details about talking face generation, fake talking face videos detection, attention mechanism and multi-modality learning that tightly related to this study. Sec. 3 introduces the proposed methodology, and Sec. 4 presents the experiments with discussions, as well as the datasets established for training and testing. In Sec. 5, we conduct ablation studies for more convincible evaluation, and conclude the whole in Sec. 6.

%% file: sec_ralated_work.tex
\section{related work}

\begin{figure}[t]
\centering
\centerline{\includegraphics [width=0.45\textwidth]{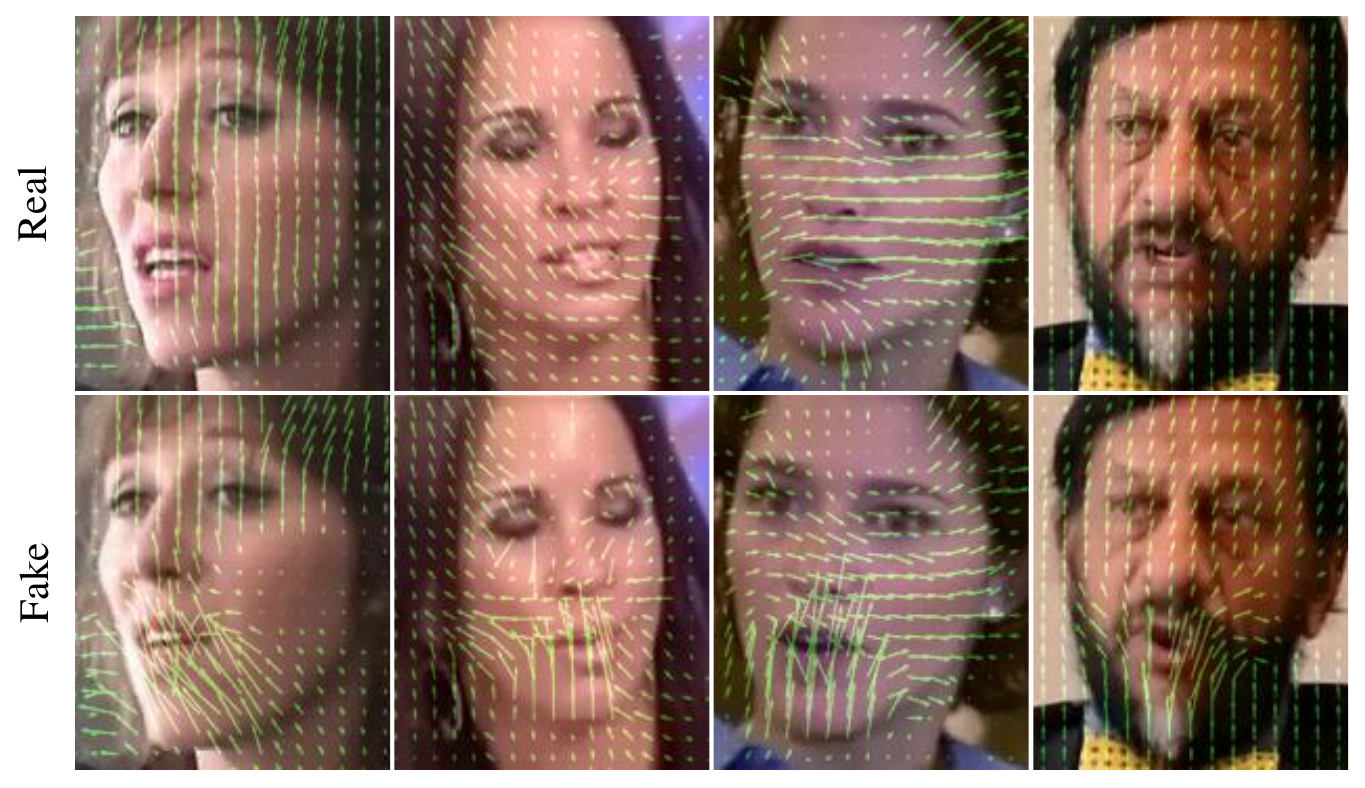}}
\setlength{\abovecaptionskip}{0.cm}
\caption{Optical flow of real video frames (top) and fake talking face video frames (bottom) estimated with Gunnar Farneback' algorithm \cite{Farnebck2003TwoFrameME}. There are obvious differences between the optical flow of real videos and fake videos, especially in the lip region.}
\setlength{\abovecaptionskip}{0.cm}
\label{img_flow}
\end{figure}

\noindent \textbf{Talking Face Generation. }To generate talking face videos from a given speech is a long-standing matter of great concern in multimedia applications. Kumar \emph{et al.} \cite{17} proposed ObamaNet, in which they employ the Char2Wav architecture (Sotelo \emph{et al.} \cite{24}) to generate speech from the input text, and then train the speech synthesis system using the audio and frames extracted from the videos. This approach can be utilized to generate lip shapes with specified identity for any text. Similarly, Suwajanakorn \emph{et al.} \cite{18} firstly maps the audio feature to sparse shape coefficients by RNNs, then maps the sparse shape to mouth texture/shapes, finally synthesizes highly detailed face textures. Jamaludin \emph{et al.} \cite{25} proposed Speech2Vid based on a joint embedding of the face and audio to obtain the synthesized talking face video frames with an encoder-decoder convolutional neural network (CNN). The GAN-based LipGAN \cite{19} model inputs the bottom-half masked target face to act as a pose prior, which guarantees that the generated mouth crops can seamlessly be fitted back into the original video without further post-processing. As an extension of LipGAN, Wav2Lip \cite{26} employs a pre-trained lip-syncing discriminator to correct the lip-syncing and a visual quality discriminator to improve the visual quality. MakeItTalk \cite{40} animates the portrait in a speaker-aware fashion driven by disentangled content and speaker embeddings. Zhou \emph{et al.} \cite{28} proposed Pose-Controllable Audio-Visual System by devising an implicit low-dimension pose code, which generates accurately lip-synced talking faces whose appearances are controllable by other videos.\\


\noindent \textbf{DeepFake Video Detection.} With rapid development with different purposes, it is hard for DeepFake techniques to avoid a dark side, such as generating fake malicious videos of celebrities and masses, which has stimulated enormous research of DeepFake video detection. Afchar \emph{et al.} \cite{20} point out that most image based fake detection can not be directly extended to video forgery detection due to the degradation of the frames by video compression. Thus, they proposed a facial video forgery detection network (MesoNet), which is integrated by CNNs with a small number of layers. Another fake detection \cite{21} is proposed by incorporating the attention mechanism into the EfficientNet \cite{29} and using the Siamese training strategy. Considering the temporal information between consecutive video frames, Tariq \emph{et al.} \cite{22} proposed CLRNet based on Convolutional LSTM and Residual Network. Sabir \emph{et al.} \cite{Sabir2019RecurrentCS} design the face manipulation detection model by combining recurrent convolutional architecture and face alignment. Haliassos \emph{et al.} \cite{Haliassos2020LipsDL} proposed LipForensics, which learns the representation of lip movements and uses the inconsistency in semantically mouth movements to achieve face forgery detection. Zhou \emph{et al.} \cite{Zhou2021FaceFI} introduced a more complex multi-person face forgery detection, constructed FFIW dataset and proposed a novel detection algorithm using multiple instance learning. Unfortunately, all of those detection strategies failed to take the audio information into account, whereas it is an essential part of the video. The ignorance of the audio information is because of none-audio-guidance to the traditional face manipulation (e.g. entire synthesis,  attribute manipulation, etc.). \\


\noindent \textbf{Fake Talking Face Video Detection.} According to the level of manipulation, the DeepFake methods are categorized into four different groups \cite{01}: (1) entire face synthesis: create entire non-existent face images using GAN models. (2) identity swap: replace the face of one person in a video with another one. (3) attribute manipulation: edit face or retouch face, consists of modifying some facial attributes such as the hair or skin colors, gender, age, glass wearing, etc. (4) expression swap: modify the facial expression of the person. In general, all of the DeepFake methods have more or fewer changes in facial features of identity. Different from DeepFake, talking face generation aims at syncing lips to match the input audio, and does not change facial identity features as shown in Fig. \ref{img_visual}. To benefit many applications such as lip-syncing for movie dubbing and lecture translation, it also enables people to produce fake talking face videos with malicious purposes, e.g. spreading fake news or extorting. In addition, the change of mouth shape usually frequently occurs in videos and leaves no trace on the person's identity, which also results in the difficulty of video forgery detection. Only by the facial features, people have little chance to discriminate whether such a video is true or not, which motivates this study of fake talking face video detection from the existing DeepFake video detection. In comparison of traditional DeepFake video detection, the more advanced talking face forgery is guided by the given speech, which means that the audio-visual representation based modeling would be a promising way to design more effective fake talking face video detectors.\\


\begin{figure*}[ht]
\centering
\centerline{\includegraphics [width=1\textwidth]{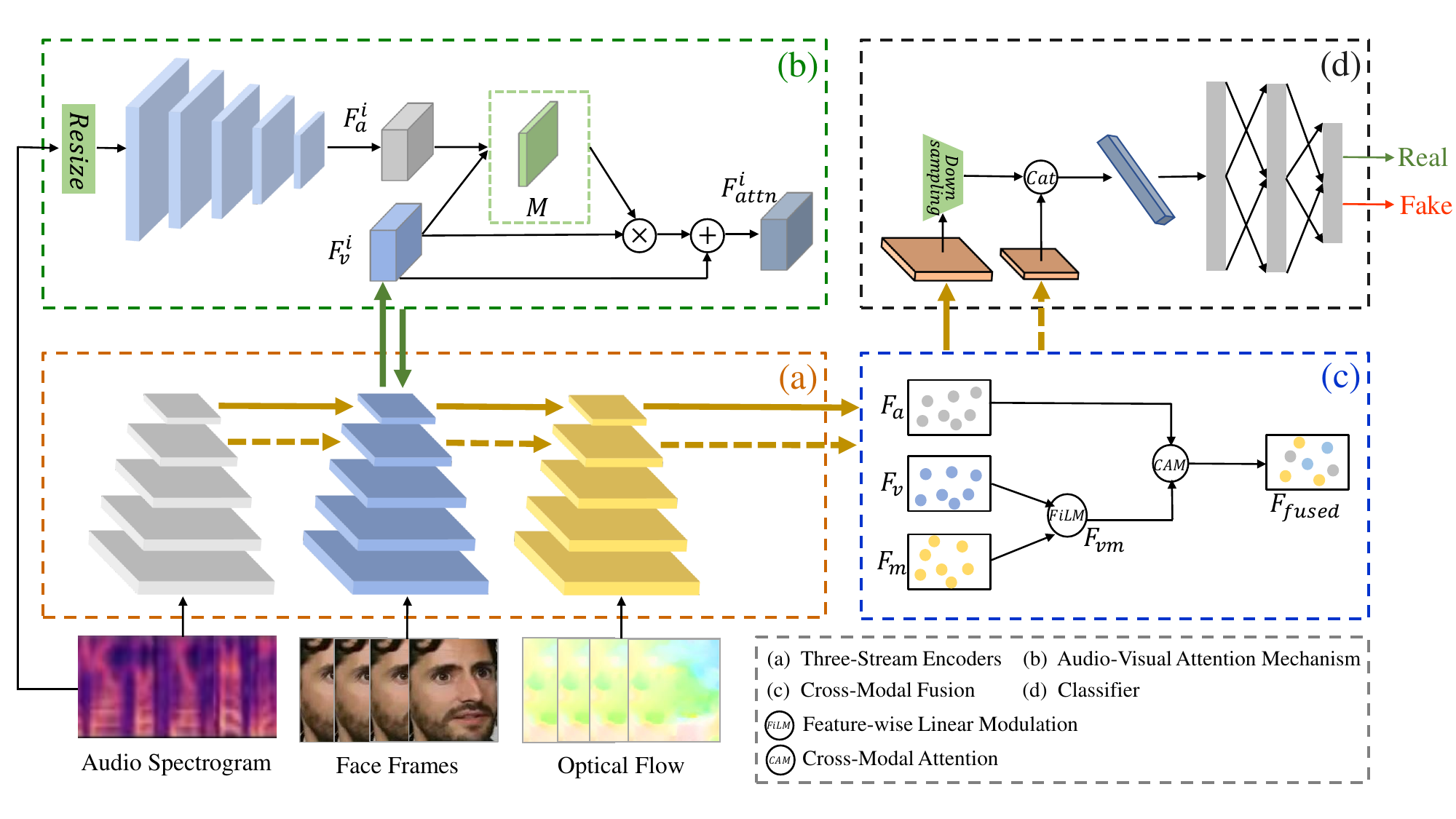}}
\setlength{\abovecaptionskip}{0.cm}
\caption{An overview of the proposed FTFDNet. FTFDNet employs three-stream encoders to learn features of the audio, visual and motion information, then uses the CMF to fuse them. Finally, a classifier is used to obtain the prediction results. Furthermore, the AVAM is embedded into each block of the visual encoder for further improvement of detection performance.}
\label{img_net}
\end{figure*}

\noindent \textbf{Audio-Visual based DeepFake Video Detection. }Audio and visual modalities are regarded to be complementary in videos, which has enlightened increasing studies on audio-visual based DeepFake video detection. By analyzing the similarity between audio and visual modalities from the video, Mittal \emph{et al.}  \cite{Mittal2020EmotionsDL} proposed a Siamese network-based architecture to learn the differences between real and fake videos. Chugh \emph{et al.}  \cite{Chugh2020NotMF} designed the modality dissonance score (MDS) to represent the similarity between audio and visual streams for the video, and further to judge whether the video is real or fake with a threshold. Similar in jointly modeling video and audio modalities, Zhou \emph{et al.} \cite{Zhou2021JointAD} proposed a detection method for the case that it is unknown whether the visual or the audio has been manipulated or not. However, the audio and visual streams are highly synchronized in fake talking face videos. Especially when the videos are generated by Wav2Lip \cite{26}, which is a network structure using SyncNet \cite{Chung2016OutOT} to correct lip-syncing, the accurate synchronization between the audio and visual would lead to the failure of detection. Therefore, the models above are  inapplicable to fake talking face video detection due to their direct-usage limitations.\\

\noindent \textbf{Multi-Modality Learning. }The trend from single-modality learning to multi-modality learning is an attractive topic in recent studies. With the capability of overcoming the limitations in perception tasks based on single-modal, audio-visual fusion has advocated more interesting works, e.g. audio-visual generation \cite{Davis2014TheVM, Duarte2019Wav2PixSF}, separation \cite{Gabbay2017SeeingTN, 9507516}, localization \cite{Tian2018AudioVisualEL, 8290832}, etc. Unfortunately, the effective fusion of different modalities is still challenging. Concatenation and element-wise sum are the most commonly used fusion operations owing to their simplicity, but the neglected internal relevance between different modalities \cite{Lai2022MultiFeatureFB} would cause the failure of joint feature representation from multiple modalities. To bridge the inter-relationship across modalities, a cross-modal fusion strategy is introduced based on the query-key-value mechanism, which is benefitted from the architecture of multi-head attention in transformer \cite{Vaswani2017AttentionIA}.\\




\noindent \textbf{Attention Mechanism. }Attention plays an important role in human perception, which enables a person to focus selectively on the salient targets instead of aimlessly processing a whole scenario at one moment. This human biological skill of visual observation can guarantee high efficiency and accuracy of our visual sensing capability, which has led to wide studies on attention mechanism \cite{46,47,48,49, 9439827, 9393325}. Typically, Woo \emph{et al.} \cite{30} proposed a convolutional block attention module (CBAM) to improve the representative ability of CNNs. By sequentially inferring attention maps along channel and spatial axes, the attention maps are multiplied with the input feature map to obtain the processed maps in which the key regions are effectively emphasized. Considering that human attention is formed by a variety of factors, e.g. when you enjoy a concert, the visual signals make you focus on the stage unconsciously, and the music leads you to stare at the salient instruments or performers, this means that visual information can be supplemented with audio signals to build up the region-of-interest. Motivated by such an interactive mechanism, an audio-visual attention mechanism is designed in this work to achieve a preferable performance in comparison to the visual-only attention mechanism. \\



%% file: sec_proposed_method.tex
\section{proposed methods}

\subsection{Framework Overview}
The proposed FTFDNet is based on multi-modal architecture as shown in Fig. \ref{img_net}. The whole process starts from the input face frames, its corresponding audio spectrogram and motion feature by three-stream encoders. The obtained high-dimensional representations of three modalities are employed to learn a unified representation by the proposed CMF efficiently. The fusion output is finally mapped into the probability to determine whether it is real or fake by the classifier with three fully connected layers, in which Dropout \cite{32} is employed to enhance robustness. Into each block of the visual encoder, the AVAM is incorporated to further improve the detection performance. The training of FTFDNet is to minimize LogLoss between the predictive probability $\hat y$ and the target label $y$ as:


\begin{equation}
L_{L} = -\frac{1}{N} \sum_{i=1}^{N}{[y_i \times \log(S(\hat y_i)) + [1 - y_i] \times \log(1 - S(\hat y_i))]}, 
\label{L_{BCE}}
\end{equation}

\noindent where $\hat y_i$ represents the predictive $i$-th face score, $y_i \in \lbrace 0,1\rbrace$ is the ground truth, and label $0$ is associated with faces coming from real pristine videos and label $1$ with fake videos. $N$ is the total number of face frames used for training and $S (\cdot)$ is the Sigmoid function.


\subsection{Three-Stream Encoders}
To achieve fake talking face video detection by the fusion of audio, visual and motion modalities, three backbone networks are employed for feature extraction. Especially, the structure of the three encoders needs to be as similar as possible, which enables the further process of intermediate features.

\noindent \textbf{Visual Encoder}. The visual backbone network is based on VGG \cite{Simonyan2014VeryDC} to encode the consecutive face frames into a high-dimensional representation. For adequate learning of temporal information between consecutive video frames, $T$ frames with the size of $T \times H \times W \times C$ are firstly reshaped to $(T \times C) \times H \times W$, then they are fed into the visual encoder. In addition, Batch Normalization \cite{31} is also incorporated to further benefit the performance of detection.


\noindent \textbf{Motion Encoder. }The motion feature is obtained from the optical flow by the motion backbone network. The structure of the motion encoder is similar to the visual encoder, and some parameter configuration of convolution layers has been changed to adapt to the reshaped size $(T \times 2) \times H \times W$ of the optical flow feature.

\noindent \textbf{Audio Encoder. }For the audio stream represented as MFCC feature, the audio backbone network is also designed based on VGG. By taking consideration of the small size of the audio spectrogram, the size reduction of the feature map is achieved by setting the parameters of convolution layers instead of the max-pooling operation.


\subsection{Cross-Modal Fusion}

\begin{figure}
  \includegraphics[width=0.48\textwidth]{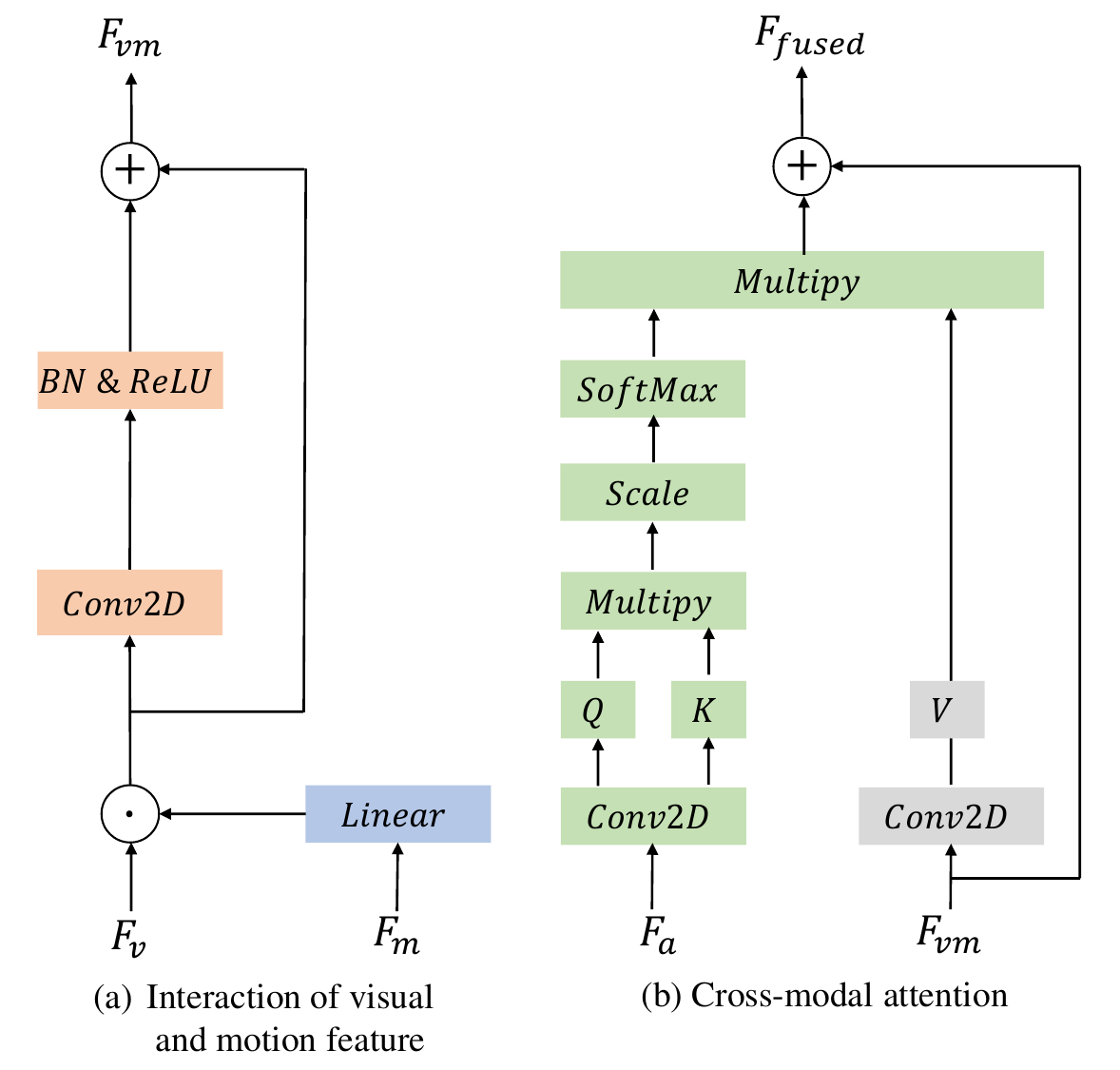}
  \caption{The structure diagram of each part in the fusion module. (a): The interaction includes affine transformation. (b): The layer of cross-modal attention.}
  \label{CMF_pic}
\end{figure}

DeepFake video detection can benefit from the synchronization between the audio and visual, which has motivated several studies \cite{Mittal2020EmotionsDL, Chugh2020NotMF, Zhou2021JointAD}. Traditional multi-modal fusion strategies (e.g. concatenation and element-wise sum) fail to take the interrelation among different modalities into consideration, especially the correlation between audio and visual. Therefore, CMF is proposed to generate the final joint representation of audio, visual and motion features.




In the Eq.\ref{film}, the FiLM \cite{Perez2017FiLMVR} is employed to perform affine transformation on visual feature ${F_v}$ by the condition of motion feature ${F_m}$. Then, a 2D convolution layer with Batch Normalization and ReLU activation is subsequently performed on the outputs of FilM to obtain modulated visual feature ${F_{vm}}$. With operations above, the modulated visual feature for joint representation learning is generated as shown in Fig. \ref{CMF_pic}(a).


  \begin{equation}
        F_{vm} = FiLM(F_m, F_v) = \gamma(F_m) \cdot F_v + \beta(F_m), 
        \label{film}
  \end{equation}

\noindent where ${\gamma(\cdot)}$ and ${\beta(\cdot)}$ are both single fully connected layer which output the scaling vector and bias vector.

Inspired by the architecture of self-attention in transformer \cite{Vaswani2017AttentionIA}, the \textbf{C}ross-\textbf{M}odal \textbf{A}ttention (CMA) is the most important process to discover the inter-relationships across modalities in the proposed fusion network, as shown in Fig. \ref{CMF_pic}(b). Given audio feature $F_{a}$ and the modulated visual feature ${F_{vm}}$, 2D convolution operations are used to generate the query ($Q$), key ($K$) and value ($V$) as Eq. \ref{kqv}. Then the fused feature $F_{fused}$ is calculated by Eq. \ref{CMA}.


\vspace{-0.1cm}
\begin{equation}
K = Conv_k(F_{a}), Q = Conv_q(F_a), V = Conv_v(F_{vm}). 
\label{kqv}
\end{equation}


  \begin{equation}
    F_{fused} = CMA(F_{vm}, F_a) = SoftMax(\frac{KQ^T}{\sqrt{d}})V + F_{vm}, 
    \label{CMA}
 \end{equation}

\noindent where $d$ denotes the dimension of $Q$, $K$ and $V$.

To substantially exploit the features from different depths of the network, a multi-scale features fusion strategy is developed: (i) Using CMF to fuse the feature maps from the last two blocks of the feature extraction network, respectively; (ii) Performing downsampling on the fused feature of the penultimate block for dimension alignment; (iii) Concatenating two fused feature representations along channel axis, and inputting the concatenated feature into the classifier to generate the probability of real and fake.


\begin{figure*}[ht]
\centering
\centerline{\includegraphics [width=0.9\textwidth]{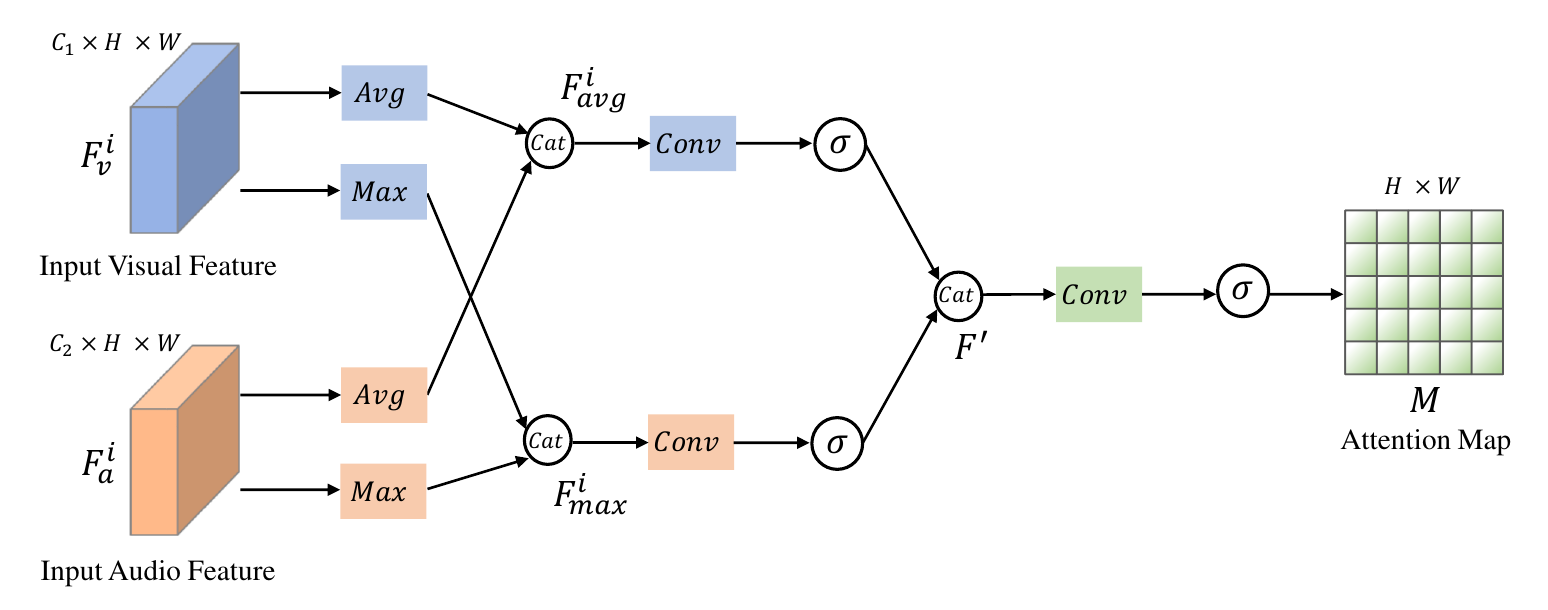}}
\setlength{\abovecaptionskip}{0.cm}
\caption{Diagram of the proposed AVAM. Compared with the conventional visual attention mechanism CBAM\cite{30}, the audio information as a supplement is integrated into the attention module in a similar way to the process of the visual feature.}
\label{img_attn}
\end{figure*}

\subsection{Audio-Visual Attention Mechanism}

The visual attention mechanism has been extensively studied \cite{30, Han2006UnsupervisedEO, Lin2017VisualAttentionBasedBM, Liu2021DeepLB}, which can be incorporated into CNN architectures for feature refinement. As the forming mechanism of human attention introduced in Sec. 2, the region-of-interest is established based on the interactions between audio and visual. Therefore, an effective audio-visual based attention mechanism is further designed to enable the detection network to focus more on the significant artifacts in the feature maps.




Given an intermediate visual feature map $F_v^i \in \mathbb{R}^{C_1 \times H \times W }$ and a corresponding audio feature map $F_a^i \in \mathbb{R}^{C_2 \times H \times W}$ as input, AVAM infers a 2D attention map $M \in \mathbb{R}^{H \times W}$ as illustrated in Fig. \ref{img_attn}. In the beginning, an average-pooling operation is applied along the channel axis to both $F_v^i$ and $F_a^i$, and then the results are concatenated to generate a feature descriptor $F_{avg}^i \in \mathbb{R}^{2 \times H \times W}$. In the same way, $F_{max}^i \in \mathbb{R}^{2 \times H \times W}$ is obtained by utilizing max-pooling operation. The above operations are shown as Eq. \ref{F_am}.

\begin{equation}
\begin{aligned}
& F_{avg}^i = Concat(AvgPool(F_v^i), AvgPool(F_a^i)),  \\
& F_{max}^i = Concat(MaxPool(F_v^i), MaxPool(F_a^i)), 
\label{F_am}
\end{aligned}
\end{equation}

\noindent where $AvgPool$ and $MaxPool$ denote the average-pooling and max-pooling operations, respectively. $Concat$ is the concatenation operation.

Then, we apply a 2D convolution layer followed by a sigmoid activation function to $F_{avg}^i$ and $F_{max}^i$ separately, and the results are concatenated to generate an intermediate feature map  $F^{\prime} \in \mathbb{R}^{2 \times H \times W}$. Finally, another 2D convolution layer followed by a sigmoid activation function is utilized to generate an attention map $M \in \mathbb{R}^{H \times W}$, which encodes the regions to be emphasized or suppressed. The above operations are shown as Eq. \ref{M}.

\begin{equation}
\begin{aligned}
& F^{\prime} = Concat(\sigma(Conv(F_{avg}^i)), \sigma(Conv(F_{max}^i))),  \\
& M = \sigma(Conv(F^{\prime})), 
\label{M}
\end{aligned}
\end{equation}

\noindent where $Concat$ is the concatenation operation, $Conv$ represents the 2D convolution operation with the filter size of $7 \times 7$, and $\sigma$ denotes the sigmoid function.

Finally, the attention map $M \in \mathbb{R}^{H \times W}$ is multiplied with the input visual feature map  $F_v^i \in \mathbb{R}^{C_1 \times H \times W }$ for adaptive feature refinement as described in Eq. \ref{AVAM}. In the new visual feature map $F_{attn}^i \in \mathbb{R}^{C_1 \times H \times W}$, the informative regions are emphasized, and the inessential parts are suppressed as well.


\begin{equation}
F_{attn}^i = F_v^i \,\otimes\, M, 
\label{AVAM}
\end{equation}

\noindent where $\otimes$ denote element-wise multiplication.


In talking face generation, the tampering usually happens in limited regions of the whole face, which indicates that the detection network should pay more attention to these informative regions, e.g. the lip region. Considering the guidance of audio to lip shape, the AVAM, instead of the conventional visual attention module, is integrated into our detection network.


The features learned from the CNN blocks of the audio encoder and visual encoder are dimensional mismatching, which can not fulfil the same size requirement of the input audio and visual feature maps in our AVAM. And considering the flexibility of integrating AVAM into other CNN architectures, in our work, the spectrogram of the input audio stream is first resized to the same size as the input visual feature. In correspondence to the visual feature map $F_v^i$, the resized audio spectrogram is input into a Siamese like visual encoder to generate the intermediate audio feature map $F_a^i$. Then by replacing the original visual feature map $F_a^i$ with the newly attentional feature map $F_{attn}^i$ computed by AVAM, a modified fake talking face detection network is formed. And the proposed AVAM is integrated into each block of the visual encoder (5 blocks) in FTFDNet to achieve overall network optimization.




%% file: sec_experiment.tex
\begin{table*}[t]
\caption{The Summary of Popular DeepFake Video Detection Datasets and Our FTFDD. Note that only two datasets among DeepFake video detection datasets and our FTFDD can be used for the research of audio-visual fusion. Our proposed FTFDD has a considerable number of videos and contains both video and audio.}
$$
\setlength{\arraycolsep}{1mm}{
\begin{array}{lcccccccc}
\toprule[1.5pt]
\rule{0pt}{10pt}
& \multicolumn{2}{c} {\text {    Videos   }} & & \multicolumn{2}{c} {\text {   Video Source   }} & & \multicolumn{2}{c} {\text {   Modes   }} \\
\cline { 2 - 3 } \cline { 5 - 6 }  \cline{ 8 - 9}
\rule{0pt}{10pt}
\text {Dataset\qquad \qquad \qquad} & \text {\quad \, Real \quad \,} & \text {\quad \, Fake \quad \,} & \text{\quad} & \text {\qquad Real \qquad} & \text {\qquad Fake \qquad} & \text{\quad} & \text{\quad Visual \quad} & \text{\quad Audio \quad}\\
\hline
\rule{0pt}{10pt}
\text {UADFV \cite{Li2018InIO} } & 49 & 49 & & \text{YouTube} & \text{FakeApp} & & $\Checkmark$ & $\XSolidBrush$\\
\rule{0pt}{10pt}
\text {DF-TIMIT \cite{Korshunov2018DeepFakesAN} } & - & 620 & & \text{\qquad VidTIMIT \cite{Sanderson2002TheVD} \qquad} & \text{faceswap-GAN} & & $\Checkmark$ & $\Checkmark$\\
\rule{0pt}{10pt}
\text {Face Forensics++ \cite{Rssler2019FaceForensicsLT} \qquad} & 1000 & 4000 & & \text{YouTube} & \text{FaceSwap \& DeepFake} & & $\Checkmark$ & $\XSolidBrush$\\
\rule{0pt}{10pt}
\text {CelebDF \cite{Li2019CelebDFAL} } & 890 & 5639 & & \text{YouTube} & \text{DeepFake} & & $\Checkmark$ & $\XSolidBrush$\\
\rule{0pt}{10pt}
\text {DFDC \cite{Dolhansky2019TheDD} } & 19154 & 99992 & & \text{Actors} & \text{Unknown} & & $\Checkmark$ & $\Checkmark$\\
\rule{0pt}{10pt}
\textbf {FTFDD(ours)} & \textbf{30000} & \textbf{34679} & & \textbf{YouTube} & \textbf{Talking Face  Generation} & & $\Checkmark$ & $\Checkmark$ \\
\bottomrule[1.5pt]
\end{array}
}
$$
\label{Table_datase_compare}
\end{table*}

\section{experiments}

\subsection{Fake Talking Face Detection Dataset}
Even the DeepFake video detection has attracted a lot of attention recently, to our limited knowledge, fake talking face video detection has not been widely studied yet. This means that the datasets of fake talking face videos are still far from sufficiency for the public. Fortunately, a series of talking face generation approaches \cite{17, 18, 19, 24, 25, 26, 27, 28, 40}  make it possible for the dataset generation. Some of the early methods are limited by identity and language, which makes them unsuitable for generating negative samples. Considering the algorithm performance and whether the code is available, in our experiment, high-performance and mainstream models (e.g. Wav2Lip \cite{26}, MakeItTalk \cite{40}, PC-AVS \cite{28}, etc.) have been employed to generate fake talking face videos on VoxCeleb2. VoxCeleb2 is an audio-visual dataset consisting of short clips of human speech from YouTube. The establishment of FTFDD is introduced in detail as below.

%
%
%

\begin{table}[h]
\centering
\caption{The Statistics of the created FTFDD. The duration, number and the lip-syncing confidence score of real, fake and total videos on FTFDD are shown in this table.}
\begin{tabular}{lccc}
\toprule[1.5pt]
\rule{0pt}{10pt}
 & \text { \quad \, Fake \quad \,} & \text { \quad \, Real \quad \,} & \text {\quad \, Total \quad \,}\\
\hline
\rule{0pt}{10pt}
\text {Total Duration(h) } & 62.28 & 61.05 & 123.33\\
\rule{0pt}{10pt}
\text {Number } & 34679 & 30000 & 64679\\
\rule{0pt}{10pt}
\text {Minimum Duration(s) } & 1.68 & 1.80 & 1.68\\
\rule{0pt}{10pt}
\text {Maximum Duration(s) } & 122.26 & 133.64 & 133.64\\
\rule{0pt}{10pt}
\text {$\rm{Sync}_{conf}$ \cite{Chung2016OutOT}} & 6.82 & 6.70 & 6.76\\
\bottomrule[1.5pt]
\end{tabular}
\label{table_dataset}
\end{table}

There are more than 6000 celebrities in VoxCeleb2 covering over 1 million utterances. Considering that each person's video in the dataset has many, only 70000 videos are selected as an alternative and any of them are randomly selected from persons' utterances. 30000 videos are chosen from the alternative as positive samples, and the rest are used to generate fake talking videos. For each video used to generate fake video, an audio stream is taken from another randomly-sampled video in the alternative. It is noteworthy that some methods of talking face generation only require a single portrait image as an identity reference (MakeItTalk and PC-AVS), which means that the first frame of the video is selected by default as the portrait image. Specifically, there is an extra video required by the PC-AVS as head pose reference, and we used the original video as the pose video reference to obtain more harmonious visions. By removing the videos of generation failure or no face detected while synthesizing, more than 30000 fake talking face videos are left. Together with the 30000 genuine videos discussed above, there are 64679 videos (each segment is at least 1.6 seconds long) included in FTFDD. The duration of FTFDD is more than 120 hours in total. And 60\% of the data is used for model training, and 20\% for model validation, 20\% for well-trained model testing. The accurate number of the videos in each class of the FTFDD can be found in Table \ref{table_dataset}. Table \ref{table_dataset} also shows the confidence score ($\rm{Sync}_{conf}$) proposed in SyncNet \cite{Chung2016OutOT} to account for the accuracy of lip-syncing. It is obvious that the confidence score of fake videos even outperforms the real videos', which proves that the fake talking face videos in FTFDD are nearly comparable to the realistic videos. Fig. \ref{img_visual} shows some sample fake faces from our FTFDD (top), DeepFake video detection datasets (bottom) including Face Forensics++ (FF++) \cite{Rssler2019FaceForensicsLT} and Celeb-DF \cite{Li2019CelebDFAL}. The fake faces in FF++ and Celeb-DF show noticeable artifacts, and our forged talking faces are indistinguishable from the real.

Table \ref{Table_datase_compare} illustrates the popular datasets of DeepFake video detection. Only DFDC \cite{Dolhansky2019TheDD} and DFTIMIT \cite{Korshunov2018DeepFakesAN} contain both audio and video, which can be used for audio-visual studies. Unlike others containing videos with manipulated faces, DFDC contains a mix of videos with manipulated faces, audio or both. The established FTFDD is also illustrated in Table \ref{Table_datase_compare}, in comparison to the other datasets, FTFDD has a large number of videos and contains both audio and visual modality, which would encourage further research in the area of audio-visual DeepFake video detection.


\subsection{Experiment Configurations}
To take full advantage of temporal information between consecutive video frames, in our experiments, we randomly choose $T=4$ consecutive frames from each video as the input of our network, and a further discussion on $T$ is conducted in ablation studies. Instead of the video frames, only the cropped face patches are input into the network because of the occurrence of mouth shape forgery during talking face generation. The face detection is performed using S3FD \cite{34}, and the obtained face crops are resized to $112 \times 112 \times 3$. The models are trained using Adam \cite{35} optimizer with default parameters ($\beta_1= 0.9$, $\beta_2= 0.999$), and initial learning rate equals to $0.001$. All experiments are conducted on 2 RTX-2080Ti GPUs, with the batch size of 32. For the evaluation, 25 continuous frames (1-second segment) from each video in the test set are used to evaluate trained models with the metrics of detection accuracy (ACC), area under the curve (AUC) and LogLoss.

\subsection{Comparison Works}
To demonstrate the superiority of the proposed multi-modal detection, comparative experiments have been conducted with the other state-of-the-art works: (1) MesoInception \cite{20}, which employs CNNs with a small number of layers based on inception architecture \cite{Szegedy2014GoingDW}. (2) Xception \cite{Rssler2019FaceForensicsLT}, a single-frame based detector which is upgraded from the Xception network. (3) LipForensics \cite{Haliassos2020LipsDL}, a multi-frame based detector by learning the inconsistencies of mouth movements. (4) CNN-GRU \cite{Sabir2019RecurrentCS}, which is implemented by ensembling DenseNet121 and GRU.

\begin{table*}[t]
\centering
\caption{Comparison of our method with other techniques on DFDC\cite{Dolhansky2019TheDD} and DFTIMIT\cite{Korshunov2018DeepFakesAN} datasets using the ACC, AUC and LogLoss metrics. Our proposed FTFDNet outperforms other visual-only and audio\_visual based DeepFake detection methods.}
\begin{tabular}{lccccccccccccc}
\toprule[1.5pt]
\rule{0pt}{10pt}
& & \multicolumn{3}{c} {\text { \qquad   DF-TIMIT(LQ) \qquad  }} & \quad & \multicolumn{3}{c} {\text { \qquad  DF-TIMIT(HQ) \qquad  }} & \quad & \multicolumn{3}{c} {\text { \qquad  DFDC \qquad  }} \\
\cline { 3 - 5 } \cline { 7 - 9 }  \cline{ 11 - 13}
\rule{0pt}{10pt}
\text {Method \qquad \qquad} & \text{\, Modality \,} & \text {\, ACC \,} & \text {\, AUC \,} & \text {LogLoss} & \text{\quad} & \text {\, ACC \,} & \text {\, AUC \,} & \text {LogLoss} & \text{\quad} & \text {\, ACC \,} & \text {\, AUC \,} & \text {LogLoss} \\
\hline
\rule{0pt}{10pt}
\text {MesoInception\cite{20}} & \text{V} & 99.36 & 99.98 & 0.3219 & & 90.12 & 97.95 & 0.4097 & & 77.68 & 78.01 & 0.5218\\
\rule{0pt}{10pt}
\text {Xception\cite{Rssler2019FaceForensicsLT}} & \text{V} & 99.04 & 99.97 & 0.0490 & & 96.67 & 99.57 & 0.1563 & & 86.41 & 90.83 & \textbf{0.3653}\\
\rule{0pt}{10pt}
\text {LipForensics\cite{Haliassos2020LipsDL}} & \text{V} & 99.33 & \textbf{100.00} & 0.0188 & & 96.00 & 99.43 & 0.3331 & & 59.99 & 56.78 & 2.0017\\
\rule{0pt}{10pt}
\textbf{CNN-GRU\cite{Sabir2019RecurrentCS}} & \text{V} & \textbf{99.73} & \textbf{100.00} & \textbf{0.0149} & & \textbf{99.07} & 99.95 & 0.0368 & & 86.64 & 91.15 & 0.4958\\
\rule{0pt}{10pt}
\text {Siamese-based\cite{Mittal2020EmotionsDL}} & \text{AV} & - & 96.30 & - & & - & 94.90 & - & & - & 84.40 & -\\
\rule{0pt}{10pt}
\text {MDS-based\cite{Chugh2020NotMF}} & \text{AV} & - & 97.90 & - & & - & 96.80 & - & & - & 90.55 & -\\
\rule{0pt}{10pt}
\textbf{FTFDNet(ours) } & \text{AV} & 99.44 & 99.99 & 0.0239 & & 99.00 & \textbf{99.97} & \textbf{0.0255} & & \textbf{88.45} & \textbf{93.35} & 0.4679\\
\bottomrule[1.5pt]
\end{tabular}
\label{Table_deepfake}
\end{table*}

\subsection{Evaluation on Fake Talking Face Detection Dataset}

\subsubsection{Quantitative Analysis}
The proposed FTFDNet and other comparison models are trained with the training set of FTFDD, and the evaluation is performed on the test set. All the detection results are shown in Table \ref{table_results}. It can be seen that the proposed model achieves significant performance superior to all the other approaches. MesoInception simply stacked with convolution layers has the worst performance, and Xception which is widely used in DeepFake detection is much better than MesoInception. LipForensics, which only learns the motion features of lips, obtains a relatively satisfactory result in this task because the forgery mainly occurs in the lip region in fake talking face videos. It's noticeable that the proposed FTFDNet (based on a variant VGG) outperforms CNN-GRU (based on a more complex and efficient backbone) and achieves the highest detection accuracy of 98.27\% and the lowest LogLoss of 0.0546.

\begin{table}[t]
\centering
\caption{Comparison of our model with other DeepFake video detection methods on our proposed FTFDD dataset using the ACC, AUC and LogLoss metrics. There are no specific methods for fake talking face video detection, so that our model can only be compared with popular DeepFake video detection methods. It's obvious that our FTFDNet achieves the best detection performance.}
\begin{tabular}{lccc}
\toprule[1.5pt]
\rule{0pt}{10pt}
\text {Method } & \text {\, ACC(\%) \,} & \text {\, AUC(\%) \,} & \text {\, LogLoss \,}\\
\hline
\rule{0pt}{10pt}
\text {MesoInception\cite{20} \quad} & 85.13 & 92.84 & 0.3708\\
\rule{0pt}{10pt}
\text {Xception\cite{Rssler2019FaceForensicsLT} } & 93.32 & 98.28 & 0.1718\\
\rule{0pt}{10pt}
\text {LipForensics\cite{Haliassos2020LipsDL} } & 96.52 & 99.57 & 0.0859\\
\rule{0pt}{10pt}
\text {CNN-GRU\cite{Sabir2019RecurrentCS} } & 97.16 & 99.65 & 0.0755\\
\rule{0pt}{10pt}
\textbf{FTFDNet(ours)} & \textbf{98.27} & \textbf{99.84} & \textbf{0.0546}\\
\bottomrule[1.5pt]
\end{tabular}
\label{table_results}
\end{table}

\begin{figure}[t]
\centering
\centerline{\includegraphics [width=0.5\textwidth]{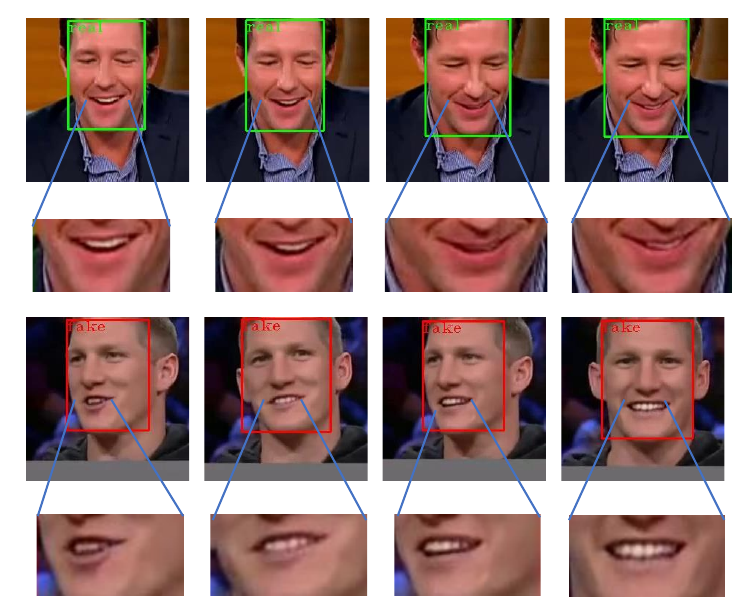}}
\setlength{\abovecaptionskip}{0.cm}
\caption{Examples of fake talking face video detection. Top line: the original talking faces from the video. Bottom line: fake talking faces from FTFDD, which are difficult to be recognized with the naked eyes. Our model can precisely identify whether any given video is real or fake.}
\label{detection_results}
\end{figure}

\subsubsection{Qualitative Analysis}

Fig. \ref{detection_results} shows the visualization of detection results of the best-performing detecting model, our FTFTNet. In Fig. \ref{detection_results}, the genuine faces are marked by a green outline (top) and the generated fake faces are marked by a red outline (bottom). By observing the enlarged mouth areas, it can be found that the fake talking faces are unidentifiable by the naked eyes, which sufficiently embodied the application value of the proposed high-precision detection model.

\begin{figure}[ht]
\centering
\centerline{\includegraphics [width=0.45\textwidth]{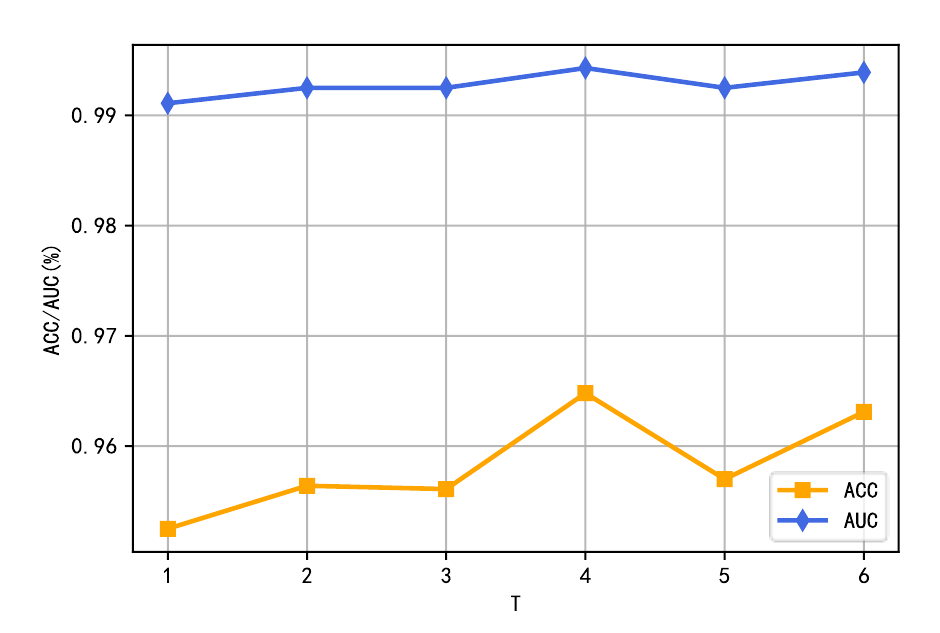}}
\setlength{\abovecaptionskip}{0.cm}
\caption{The ACC and AUC of $T$ with different values on the FTFDD dataset. It's observed that when $T=4$,  the audio-visual network performs the best.}
\label{ablation_T}
\end{figure}

\begin{figure*}[t]
\centering
\centerline{\includegraphics [width=0.9\textwidth]{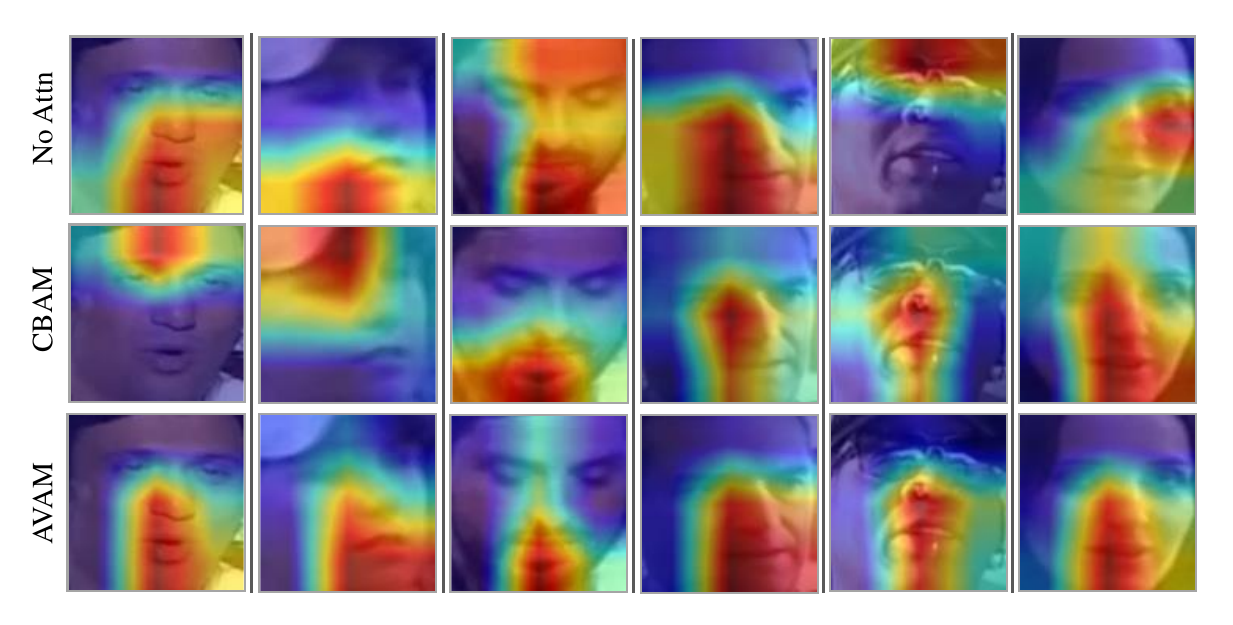}}
\setlength{\abovecaptionskip}{0.cm}
\caption{Visualization of feature maps using Grad-CAM\cite{Selvaraju2016GradCAMVE}. Some examples of final fused feature maps from w/o Attention (top), w CBAM (middle) and w AVAM (bottom) are visualized in their original face crops.}
\label{attention map}
\end{figure*}

\subsection{Evaluation on DeepFake Video Detection Datasets}

To validate the effectiveness of face forgery detection for the proposed FTFDNet, extensive experiments have also been conducted on the popular DeepFake video detection  datasets of DFDC\cite{Dolhansky2019TheDD} and DFTIMIT\cite{Korshunov2018DeepFakesAN}, which contain the modalities of both visual and audio. Specifically, the DF-TIMIT has two different forgeries: Low Quality (LQ) and High Quality (HQ). Since both of the forgeries only contain the videos with manipulated faces, the real video source VidTIMIT \cite{Sanderson2002TheVD} is then used as positive samples. For DFDC in the experiments, 18000 videos are randomly sampled in consideration of computational overhead. Both DF-TIMIT and DFDC are split into training set (60\%), validation set (20\%), and testing set (20\%) as same as the established FTFDD.


All the detection results are presented in Table \ref{Table_deepfake}. For the approaches whose implementation is not available, we directly refer the results reported in their papers, and "\_" denotes the results are unavailable. Owing to a smaller quantity of videos and obvious artifacts in fake videos, all detection methods achieve high detection performance on DF-TIMIT (LQ). Both FTFDNet and CNN-GRU also achieve high accuracy of 99\% on DF-TIMIT (HQ) and significantly outperform other methods as well. For the DFDC with complex video forgery, LipForensics shows a very poor detection performance because the learning of forgery features is limited within the lip regions. It is noteworthy that the proposed FTFDNet outperforms the other visual-only and audio-visual based detection with the accuracy of 88.45\% and AUC of 93.35\%, which demonstrates a superior performance compared to the other state-of-the-art works in not only the fake talking face detection but also the conventional DeepFake video detection.

\begin{table*}[t]
\centering
\caption{The results of MesoInception \cite{20} with different attention modules on DFDC \cite{Dolhansky2019TheDD} and DFTIMIT(HQ) \cite{Korshunov2018DeepFakesAN} datasets. It's confirmed that our AVAM can be integrated with other CNN networks and brings greater improvement of detection performance.}
\begin{tabular}{lccccccccc}
\toprule[1.5pt]
\rule{0pt}{10pt}
 & \multicolumn{3}{c} {\text { \qquad   DF-TIMIT(HQ) \qquad  }} & \qquad & \multicolumn{3}{c} {\text { \qquad  DFDC \qquad  }}\\
\cline { 2 - 4 } \cline { 6 - 8 }
\rule{0pt}{10pt}
\text {Method}  & \text {\, ACC \,} & \text {\, AUC \,} & \text {LogLoss} & \text{\qquad} & \text {\, ACC \,} & \text {\, AUC \,} & \text {LogLoss}\\
\hline
\rule{0pt}{10pt}
\text {MesoInception\cite{20}} & 90.12 & 97.95 & 0.4097 & & 77.68 & 78.01 & 0.5281\\
\rule{0pt}{10pt}
\text {MesoInception $+$ CBAM \qquad} & 91.84 & 99.07 & 0.3914 & & 78.11 & 78.72 & 0.5232\\
\rule{0pt}{10pt}
\textbf{MesoInception $+$ AVAM \qquad} & \textbf{95.39} & \textbf{99.31} & \textbf{0.3605} & & \textbf{78.81} & \textbf{81.16} & \textbf{0.5134}\\
\bottomrule[1.5pt]
\end{tabular}
\label{Table_Meso_attn}
\end{table*}

\subsection{Ablation Studies}

\subsubsection{Ablation Studies on Input Video Frame Number}

In \cite{22}, Tariq \emph{et al.} indicated that the temporal information between consecutive video frames (5 consecutive frames as the input) is effective to discover the imperceptible artifacts. Usually for available talking face generation models, consecutive frames from a pre-defined buffer $T$ ($T=5$ in Wav2Lip \cite{26}, LipGan \cite{19} and PC-AVS \cite{28}) are input to obtain the temporal context information. This also means that a strong correlation between lips can be formed during talking face generation, and further demonstrates the necessity for multi-frame exploration. Similar by using a frame buffer of 5, SyncNet \cite{Chung2016OutOT} is designed to evaluate the synchronisation between mouth motion and speech in a video. These works motivates us to design the proposed multi-frame detection network with about 5 consecutive frames for model input.



The ablation study is conducted on the number of consecutive frames, which are randomly sampled from each video on FTFDD as the input of networks. Fig. \ref{ablation_T} reports the curves of ACC and AUC with $T \in {1, 2, ..., 6}$ on the audio-visual network (audio and visual encoder without AVAM of FTFDNet). It can be found that the inter-frame information has indeed enhanced the accuracy of fake talking face detection (The accuracy of $T>1$ is higher than the accuracy of $T=1$) as expected. At the same time, it is found that a bigger buffer size does not mean better performance (The accuracy of $T=4$ is higher than the accuracy of $T=5$). The possible reason is that the accumulated temporal context information between the frames also increases the complexity of the recognition, which needs to be balanced. Through experimental verification, we set $T$ to 4 to achieve the best performance.

\begin{table}
\centering
\caption{The detection results of FTFDNet with different attention mechanisms on the FTFDD dataset. To verify the validity of audio-visual based AVAM, we compared it with the visual-only based CBAM.}
\begin{tabular}{lccc}
\toprule[1.5pt]
\rule{0pt}{10pt}
\text {Method } & \text {\quad ACC(\%) \quad} & \text {\quad AUC(\%) \quad} & \text {\quad LogLoss \quad}\\
\hline
\rule{0pt}{10pt}
\text {w/o Attention } & 96.48 & 99.43 & 0.1009\\
\rule{0pt}{10pt}
\text {w CBAM } & 96.82 & 99.53 & 0.0925\\
\rule{0pt}{10pt}
\textbf{w AVAM } & \textbf{97.31} & \textbf{99.65} & \textbf{0.0753}\\
\bottomrule[1.5pt]
\end{tabular}
\label{Table_AVAM}
\end{table}

\begin{table}[t]
\centering
\caption{The detection results of FTFDNet with different multi-modal fusion strategies on the FTFDD dataset. To verify the validity of the propsed CMF, we compared it with conventional fusion methods including concatenation and element-wise sum.}
\begin{tabular}{lccc}
\toprule[1.5pt]
\rule{0pt}{10pt}
\text {Method } & \text {\quad ACC(\%) \quad} & \text {\quad AUC(\%) \quad} & \text {\, LogLoss \,}\\
\hline
\rule{0pt}{10pt}
\text {Concatenation} & 97.97 & 99.82 & 0.0601\\
\rule{0pt}{10pt}
\text {Element-wise Sum} & 97.10  &  99.62 & 0.0804 \\
\rule{0pt}{10pt}
\text {CMF-kqv} & 97.18  &  99.62 & 0.083 \\
\rule{0pt}{10pt}
\text {CMF-last} & 97.80  &  99.76 & 0.066 \\
\rule{0pt}{10pt}
\textbf{CMF} & \textbf{98.27} & \textbf{99.84} & \textbf{0.0546}\\
\rule{0pt}{10pt}
\text {CMF-last3} & 97.63  &  99.75 & 0.0696 \\
\bottomrule[1.5pt]
\end{tabular}
\label{Table_CMF}
\end{table}

\begin{table}[t]
\centering
\caption{The detection results of integrating different modalities on the FTFDD dataset. The audio and motion features are gradually embedded into the visual-only network, and both can improve the detection performance.}
\begin{tabular}{cccccc}
\toprule[1.5pt]
\rule{0pt}{10pt}
\text{Visual} & \text{Audio} & \text{Motion} & \text{\, ACC(\%) \,} & \text{\, AUC(\%) \,} & \text{\, LogLoss}\\
\hline
\rule{0pt}{10pt}
$\surd$ & & & 95.14 & 99.03 & 0.1349\\
\rule{0pt}{10pt}
 & $\surd$ & & 62.01 & 68.71 & 0.6404\\
\rule{0pt}{10pt}
 & & $\surd$ & 86.99 & 94.61 & 0.3001\\
\rule{0pt}{10pt}
$\surd$ & $\surd$ & & 96.48 & 99.43 & 0.1009\\
\rule{0pt}{10pt}
$\bm{\surd}$ & $\bm{\surd}$ & $\bm{\surd}$ & \textbf{97.56} & \textbf{99.70} & \textbf{0.0778}\\
\bottomrule[1.5pt]
\end{tabular}
\label{Table_module}
\end{table}

\subsubsection{Ablation Studies on AVAM}

In order to confirm the performance of our proposed AVAM, we conduct ablation experiments: (1) w/o Attention, which represents the audio-visual network. (2) w CBAM, which incorporates CBAM \cite{30} into the audio-visual network in a similar way as FTFDNet. (3) w AVAM, which incorporates our proposed AVAM into the audio-visual network. As shown in Table \ref{Table_AVAM}, both CBAM and AVAM can improve the performance of the detection model, and our AVAM achieves an increase of 0.83\% on detection accuracy, which is significantly superior to the performance of  CBAM with an increase of 0.34\%. The experimental results have demonstrated that audio information is useful to build up the region of interest in audio-visual tasks.


Fig. \ref{attention map} shows the visualization of fused feature maps using Grad-CAM \cite{Selvaraju2016GradCAMVE}, where the feature maps are extracted from the last convolution block of w/o Attention (top), w CBAM (middle) and w AVAM (bottom). It is found that after integrating the attention mechanism, the network produces different degrees of attention to different regions in the whole face image. Compared to the conventional visual attention mechanism CBAM, the proposed AVAM enables the fake detection network to focus attention on the lip regions which might be tampered by talking face generation methods. This verifies the usefulness of audio information for key regions extraction. Once these informative regions have been emphasized, the fake talking face detection network is able to achieve higher detection accuracy.

Similar to CBAM, the proposed AVAM can be incorporated into more CNN architectures for audio-visual learning. To verify the effectiveness of AVAM on other CNN based networks, we add AVAM into visual-only based MesoInception\cite{20}. CBAM and AVAM are respectively incorporated into each block of  MesoInception network in the same way as FTFDNet. According to the requirement of the audio stream for AVAM, the same structure of MesoInception is employed to design an audio branch for audio features generation in correspondence to obtained visual features. The detection results on DFDC\cite{Dolhansky2019TheDD} and DFTIMIT (HQ)\cite{Korshunov2018DeepFakesAN} datasets are shown in Table \ref{Table_Meso_attn}. It's found that the MesoInception with AVAM performs the best, which has shown a capability of flexible integration of our AVAM with other CNN models. Besides, the performance improvement is also greater than the visual-only based attention mechanism CBAM.



\subsubsection{Ablation Studies on Fusion Strategies}

Owing to the simplicity and applicability, concatenation and element-wise sum are the most commonly used fusion strategies. These simple fusion methods ignore the correlation between different modalities. In contrast, the proposed CMF can effectively achieve a joint representation of different modalities in another way, which is inspired by multi-head attention architecture in transformer. To verify the superiority of CMF compared with conventional fusion strategies, ablation experiments are conducted on the FTFDD dataset by changing the fusion module in FTFDNet. The experimental results are shown in Table \ref{Table_CMF}, it can be easily found that the fusion based on CMF performs better than the common fusion methods of concatenation and element-wise sum, which indicates that the inter-relationship across different modalities can be effectively bridged by the proposed fusion.

To discuss the details of our fusion strategy set up, we conduct the experiment CMF-kqv, in which the key ($K$), value ($V$) are generated by the modulated visual feature ${F_{vm}}$ and the query ($Q$) is calculated by the audio feature $F_{a}$. The CMF-kqv underperforms the CMF, which shows that the architecture of self-attention in CMF is superior. Furthermore, Table \ref{Table_CMF} shows the detection results of fusing features from different stages in the backbone network. (CMF-last: only using the feature from the last stage.  CMF: using the feature from the last two stages. CMF-last3: using the feature from the last three stages.) The results proved that the multi-scale fusion strategy of the last two stages outperform the best.

\subsubsection{Ablation Studies on Modalities}

To verify the contribution of each modality to the detection performance, we conduct ablation experiments on audio, visual and motion modalities. Table \ref{Table_module} lists the detection results of using each modality and combination of multiple modalities. As shown in Table \ref{Table_module}, using only audio feature achieved the lowest performance because of no manipulation in audio from the video, and only motion feature based network obtained the accuracy of more than 80\%, which confirmed that there are differences in the optical flow between real and fake videos. Table \ref{Table_module} also shows a detection accuracy increase of 1.34\% when audio information is used to help the detection model capture the manipulation cues. There are differences between the optical flow of real video frames and fake video frames. To verify that the motion feature from optical flow is helpful to capture the subtle artifact cues, we integrate the audio, visual and motion features using our proposed CMF, and a detection accuracy increase of 1.08\% is obtained compared with the audio-visual network.

%% file: sec_conclusion.tex
\section{Limitation and Discussion}
Our established dataset, unfortunately, is constrained by the utilized talking face generation methods, such as the lip synchronization accuracy and the quality of the generated images produced by current algorithms. The utilization of low-resolution real talking face videos also contributes to the production of low-quality forged videos. With the advancement in talking face generation techniques, maintaining all-embracing for fake talking face datasets poses a formidable challenge.

To solve the more challenging task of fake talking face video detection, in this paper, we propose a novel \textbf{F}ake \textbf{T}alking \textbf{F}ace \textbf{D}etection \textbf{N}etwork (FTFDNet) by incorporating audio, visual and motion features using an efficient \textbf{C}ross-\textbf{M}odal \textbf{F}usion (CMF) strategy. Beyond that, we propose an \textbf{A}udio-\textbf{V}isual \textbf{A}ttention \textbf{M}echanism (AVAM), which enables the network to focus on the most relevant portions of the feature maps. To further improve the detection performance, our proposed AVAM is embedded into FTFDNet and obtained a significant performance boost over the popular visual attention mechanism. Training and evaluating of the proposed FTFDNet are performed on the established \textbf{F}ake \textbf{T}alking \textbf{F}ace \textbf{D}etection \textbf{D}ataset (FTFDD) and popular DeepFake video detection datasets, the proposed FTFDNet shows excellent performance on the detection of not only fake talking face videos but also DeepFake videos. 

%% file: New_IEEEtran_how-to.bbl
\begin{thebibliography}{10}
\providecommand{\url}[1]{#1}
\csname url@samestyle\endcsname
\providecommand{\newblock}{\relax}
\providecommand{\bibinfo}[2]{#2}
\providecommand{\BIBentrySTDinterwordspacing}{\spaceskip=0pt\relax}
\providecommand{\BIBentryALTinterwordstretchfactor}{4}
\providecommand{\BIBentryALTinterwordspacing}{\spaceskip=\fontdimen2\font plus
\BIBentryALTinterwordstretchfactor\fontdimen3\font minus
  \fontdimen4\font\relax}
\providecommand{\BIBforeignlanguage}[2]{{%
\expandafter\ifx\csname l@#1\endcsname\relax
\typeout{** WARNING: IEEEtran.bst: No hyphenation pattern has been}%
\typeout{** loaded for the language `#1'. Using the pattern for}%
\typeout{** the default language instead.}%
\else
\language=\csname l@#1\endcsname
\fi
#2}}
\providecommand{\BIBdecl}{\relax}
\BIBdecl

\bibitem{02}
T.~Karras, S.~Laine, and T.~Aila, ``A style-based generator architecture for
  generative adversarial networks,'' \emph{2019 IEEE/CVF Conference on Computer
  Vision and Pattern Recognition (CVPR)}, pp. 4396--4405, 2019.

\bibitem{03}
T.~Karras, T.~Aila, S.~Laine, and J.~Lehtinen, ``Progressive growing of gans
  for improved quality, stability, and variation,'' \emph{ArXiv}, vol.
  abs/1710.10196, 2018.

\bibitem{04}
F.~Marra, D.~Gragnaniello, L.~Verdoliva, and G.~Poggi, ``Do gans leave
  artificial fingerprints?'' \emph{2019 IEEE Conference on Multimedia
  Information Processing and Retrieval (MIPR)}, pp. 506--511, 2019.

\bibitem{06}
J.~Thies, M.~Zollh{\"o}fer, and M.~Nie{\ss}ner, ``Deferred neural rendering:
  Image synthesis using neural textures,'' \emph{arXiv: Computer Vision and
  Pattern Recognition}, 2019.

\bibitem{07}
J.~Thies, M.~Zollh{\"o}fer, M.~Stamminger, C.~Theobalt, and M.~Nie{\ss}ner,
  ``Face2face: real-time face capture and reenactment of rgb videos,''
  \emph{ArXiv}, vol. abs/2007.14808, 2019.

\bibitem{01}
R.~Tolosana, R.~Vera-Rodr{\'i}guez, J.~Fierrez, A.~Morales, and
  J.~Ortega-Garcia, ``Deepfakes and beyond: A survey of face manipulation and
  fake detection,'' \emph{Inf. Fusion}, vol.~64, pp. 131--148, 2020.

\bibitem{08}
S.~McCloskey and M.~Albright, ``Detecting gan-generated imagery using color
  cues,'' \emph{ArXiv}, vol. abs/1812.08247, 2018.

\bibitem{09}
F.~Marra, C.~Saltori, G.~Boato, and L.~Verdoliva, ``Incremental learning for
  the detection and classification of gan-generated images,'' \emph{2019 IEEE
  International Workshop on Information Forensics and Security (WIFS)}, pp.
  1--6, 2019.

\bibitem{10}
N.~Hulzebosch, S.~Ibrahimi, and M.~Worring, ``Detecting cnn-generated facial
  images in real-world scenarios,'' \emph{2020 IEEE/CVF Conference on Computer
  Vision and Pattern Recognition Workshops (CVPRW)}, pp. 2729--2738, 2020.

\bibitem{11}
S.~Agarwal, H.~Farid, Y.~Gu, M.~He, K.~Nagano, and H.~Li, ``Protecting world
  leaders against deep fakes,'' in \emph{CVPR Workshops}, 2019.

\bibitem{12}
T.~hyun Jung, S.~Kim, and K.~Kim, ``Deepvision: Deepfakes detection using human
  eye blinking pattern,'' \emph{IEEE Access}, vol.~8, pp. 83\,144--83\,154,
  2020.

\bibitem{13}
Y.~Li and S.~Lyu, ``Exposing deepfake videos by detecting face warping
  artifacts,'' \emph{ArXiv}, vol. abs/1811.00656, 2019.

\bibitem{14}
A.~R{\"o}ssler, D.~Cozzolino, L.~Verdoliva, C.~Riess, J.~Thies, and
  M.~Nie{\ss}ner, ``Faceforensics++: Learning to detect manipulated facial
  images,'' \emph{2019 IEEE/CVF International Conference on Computer Vision
  (ICCV)}, pp. 1--11, 2019.

\bibitem{15}
J.~Stehouwer, H.~Dang, F.~Liu, X.~Liu, and A.~K. Jain, ``On the detection of
  digital face manipulation,'' \emph{2020 IEEE/CVF Conference on Computer
  Vision and Pattern Recognition (CVPR)}, pp. 5780--5789, 2020.

\bibitem{16}
E.~Sabir, J.~Cheng, A.~Jaiswal, W.~AbdAlmageed, I.~Masi, and P.~Natarajan,
  ``Recurrent convolutional strategies for face manipulation detection in
  videos,'' in \emph{CVPR Workshops}, 2019.

\bibitem{17}
R.~Kumar, J.~M.~R. Sotelo, K.~Kumar, A.~D. Br{\'e}bisson, and Y.~Bengio,
  ``Obamanet: Photo-realistic lip-sync from text,'' \emph{ArXiv}, vol.
  abs/1801.01442, 2018.

\bibitem{18}
S.~Suwajanakorn, S.~Seitz, and I.~Kemelmacher-Shlizerman, ``Synthesizing
  obama,'' \emph{ACM Transactions on Graphics (TOG)}, vol.~36, pp. 1 -- 13,
  2017.

\bibitem{24}
J.~M.~R. Sotelo, S.~Mehri, K.~Kumar, J.~F. Santos, K.~Kastner, A.~C. Courville,
  and Y.~Bengio, ``Char2wav: End-to-end speech synthesis,'' in \emph{ICLR},
  2017.

\bibitem{19}
R.~PrajwalK, R.~Mukhopadhyay, J.~Philip, A.~Jha, V.~Namboodiri, and C.~V.
  Jawahar, ``Towards automatic face-to-face translation,'' \emph{Proceedings of
  the 27th ACM International Conference on Multimedia}, 2019.

\bibitem{26}
R.~PrajwalK, R.~Mukhopadhyay, V.~Namboodiri, and C.~V. Jawahar, ``A lip sync
  expert is all you need for speech to lip generation in the wild,''
  \emph{Proceedings of the 28th ACM International Conference on Multimedia},
  2020.

\bibitem{27}
G.~Wang, P.~Zhang, L.~Xie, W.~Huang, and Y.~Zha, ``Attention-based lip
  audio-visual synthesis for talking face generation in the wild,'' 2021.

\bibitem{40}
H.~Tan, Y.~Zhou, Q.~Tao, J.~Rosen, and S.~van Dijken, ``Bioinspired
  multisensory neural network with crossmodal integration and recognition,''
  \emph{Nature Communications}, vol.~12, 2021.

\bibitem{28}
H.~Zhou, Y.~Sun, W.~Wu, C.~C. Loy, X.~Wang, and Z.~Liu, ``Pose-controllable
  talking face generation by implicitly modularized audio-visual
  representation,'' \emph{ArXiv}, vol. abs/2104.11116, 2021.

\bibitem{8995571}
L.~Yu, J.~Yu, M.~Li, and Q.~Ling, ``Multimodal inputs driven talking face
  generation with spatial–temporal dependency,'' \emph{IEEE Transactions on
  Circuits and Systems for Video Technology}, vol.~31, no.~1, pp. 203--216,
  2021.

\bibitem{Rssler2019FaceForensicsLT}
A.~R{\"o}ssler, D.~Cozzolino, L.~Verdoliva, C.~Riess, J.~Thies, and
  M.~Nie{\ss}ner, ``Faceforensics++: Learning to detect manipulated facial
  images,'' \emph{2019 IEEE/CVF International Conference on Computer Vision
  (ICCV)}, pp. 1--11, 2019.

\bibitem{Li2019CelebDFAL}
Y.~Li, X.~Yang, P.~Sun, H.~Qi, and S.~Lyu, ``Celeb-df: A large-scale
  challenging dataset for deepfake forensics,'' \emph{2020 IEEE/CVF Conference
  on Computer Vision and Pattern Recognition (CVPR)}, pp. 3204--3213, 2019.

\bibitem{Farid2008ASO}
H.~Farid, ``A survey of image forgery detection,'' 2008.

\bibitem{9878273}
J.~Chen, X.~Liao, W.~Wang, Z.~Qian, Z.~Qin, and Y.~Wang, ``Snis: A signal noise
  separation-based network for post-processed image forgery detection,''
  \emph{IEEE Transactions on Circuits and Systems for Video Technology},
  vol.~33, no.~2, pp. 935--951, 2023.

\bibitem{9878360}
X.~Bi, Y.~Shang, B.~Liu, B.~Xiao, W.~Li, and X.~Gao, ``A versatile detection
  method for various contrast enhancement manipulations,'' \emph{IEEE
  Transactions on Circuits and Systems for Video Technology}, vol.~33, no.~2,
  pp. 491--504, 2023.

\bibitem{20}
D.~Afchar, V.~Nozick, J.~Yamagishi, and I.~Echizen, ``Mesonet: a compact facial
  video forgery detection network,'' \emph{2018 IEEE International Workshop on
  Information Forensics and Security (WIFS)}, pp. 1--7, 2018.

\bibitem{21}
N.~Bonettini, E.~D. Cannas, S.~Mandelli, L.~Bondi, P.~Bestagini, and S.~Tubaro,
  ``Video face manipulation detection through ensemble of cnns,'' \emph{2020
  25th International Conference on Pattern Recognition (ICPR)}, pp. 5012--5019,
  2021.

\bibitem{22}
S.~Tariq, S.~Lee, and S.~S. Woo, ``A convolutional lstm based residual network
  for deepfake video detection,'' \emph{ArXiv}, vol. abs/2009.07480, 2020.

\bibitem{23}
J.~Pu, N.~Mangaokar, L.~Kelly, P.~Bhattacharya, K.~Sundaram, M.~Javed, B.~Wang,
  and B.~Viswanath, ``Deepfake videos in the wild: Analysis and detection,''
  \emph{Proceedings of the Web Conference 2021}, 2021.

\bibitem{9643421}
J.~Yang, S.~Xiao, A.~Li, W.~Lu, X.~Gao, and Y.~Li, ``Msta-net: Forgery
  detection by generating manipulation trace based on multi-scale self-texture
  attention,'' \emph{IEEE Transactions on Circuits and Systems for Video
  Technology}, vol.~32, no.~7, pp. 4854--4866, 2022.

\bibitem{Chugh2020NotMF}
K.~Chugh, P.~Gupta, A.~Dhall, and R.~Subramanian, ``Not made for each other-
  audio-visual dissonance-based deepfake detection and localization,''
  \emph{Proceedings of the 28th ACM International Conference on Multimedia},
  2020.

\bibitem{Chung2016OutOT}
J.~S. Chung and A.~Zisserman, ``Out of time: Automated lip sync in the wild,''
  in \emph{ACCV Workshops}, 2016.

\bibitem{Zhou2021JointAD}
Y.~Zhou and S.-N. Lim, ``Joint audio-visual deepfake detection,'' \emph{2021
  IEEE/CVF International Conference on Computer Vision (ICCV)}, pp.
  14\,780--14\,789, 2021.

\bibitem{36}
M.~Wallace, M.~Meredith, and B.~Stein, ``Multisensory integration in the
  superior colliculus of the alert cat.'' \emph{Journal of neurophysiology},
  vol. 80 2, pp. 1006--10, 1998.

\bibitem{37}
A.~King, ``Multisensory integration: Strategies for synchronization,''
  \emph{Current Biology}, vol.~15, pp. R339--R341, 2005.

\bibitem{38}
T.~Ohshiro, D.~Angelaki, and G.~DeAngelis, ``A normalization model of
  multisensory integration,'' \emph{Nature neuroscience}, vol.~14, pp. 775 --
  782, 2011.

\bibitem{39}
J.~McDonald, W.~Teder-S{\"a}lej{\"a}rvi, and L.~Ward, ``Multisensory
  integration and crossmodal attention effects in the human brain.''
  \emph{Science}, vol. 292 5523, p. 1791, 2001.

\bibitem{41}
L.~Franzen, I.~Delis, G.~D. Sousa, C.~Kayser, and M.~Philiastides, ``Auditory
  information enhances post-sensory visual evidence during rapid multisensory
  decision-making,'' \emph{Nature Communications}, vol.~11, 2020.

\bibitem{Farnebck2003TwoFrameME}
G.~Farneb{\"a}ck, ``Two-frame motion estimation based on polynomial
  expansion,'' in \emph{Scandinavian Conference on Image Analysis}, 2003.

\bibitem{Dolhansky2019TheDD}
B.~Dolhansky, R.~Howes, B.~Pflaum, N.~Baram, and C.~Canton-Ferrer, ``The
  deepfake detection challenge (dfdc) preview dataset,'' \emph{ArXiv}, vol.
  abs/1910.08854, 2019.

\bibitem{Korshunov2018DeepFakesAN}
P.~Korshunov and S.~Marcel, ``Deepfakes: a new threat to face recognition?
  assessment and detection,'' \emph{ArXiv}, vol. abs/1812.08685, 2018.

\bibitem{25}
A.~Jamaludin, J.~S. Chung, and A.~Zisserman, ``You said that?: Synthesising
  talking faces from audio,'' \emph{International Journal of Computer Vision},
  pp. 1--13, 2019.

\bibitem{29}
M.~Tan and Q.~V. Le, ``Efficientnet: Rethinking model scaling for convolutional
  neural networks,'' \emph{ArXiv}, vol. abs/1905.11946, 2019.

\bibitem{Sabir2019RecurrentCS}
E.~Sabir, J.~Cheng, A.~Jaiswal, W.~AbdAlmageed, I.~Masi, and P.~Natarajan,
  ``Recurrent convolutional strategies for face manipulation detection in
  videos,'' in \emph{CVPR Workshops}, 2019.

\bibitem{Haliassos2020LipsDL}
A.~Haliassos, K.~Vougioukas, S.~Petridis, and M.~Pantic, ``Lips don't lie: A
  generalisable and robust approach to face forgery detection,'' \emph{2021
  IEEE/CVF Conference on Computer Vision and Pattern Recognition (CVPR)}, pp.
  5037--5047, 2020.

\bibitem{Zhou2021FaceFI}
T.~Zhou, W.~Wang, Z.~Liang, and J.~Shen, ``Face forensics in the wild,''
  \emph{2021 IEEE/CVF Conference on Computer Vision and Pattern Recognition
  (CVPR)}, pp. 5774--5784, 2021.

\bibitem{Mittal2020EmotionsDL}
T.~Mittal, U.~Bhattacharya, R.~Chandra, A.~Bera, and D.~Manocha, ``Emotions
  don't lie: An audio-visual deepfake detection method using affective cues,''
  \emph{Proceedings of the 28th ACM International Conference on Multimedia},
  2020.

\bibitem{Davis2014TheVM}
A.~Davis, M.~Rubinstein, N.~Wadhwa, G.~J. Mysore, F.~Durand, and W.~T. Freeman,
  ``The visual microphone: Passive recovery of sound from video,'' 2014.

\bibitem{Duarte2019Wav2PixSF}
A.~C. Duarte, F.~Roldan, M.~Tubau, J.~Escur, S.~Pascual, A.~Salvador,
  E.~Mohedano, K.~McGuinness, J.~Torres, and X.~G. i~Nieto, ``Wav2pix:
  Speech-conditioned face generation using generative adversarial networks,''
  \emph{ICASSP 2019 - 2019 IEEE International Conference on Acoustics, Speech
  and Signal Processing (ICASSP)}, pp. 8633--8637, 2019.

\bibitem{Gabbay2017SeeingTN}
A.~Gabbay, A.~Ephrat, T.~Halperin, and S.~Peleg, ``Seeing through noise:
  Visually driven speaker separation and enhancement,'' in \emph{IEEE
  International Conference on Acoustics, Speech, and Signal Processing}, 2017.

\bibitem{9507516}
Y.~He, X.~Xu, J.~Zhang, F.~Shen, Y.~Yang, and H.~T. Shen, ``Modeling two-stream
  correspondence for visual sound separation,'' \emph{IEEE Transactions on
  Circuits and Systems for Video Technology}, vol.~32, no.~5, pp. 3291--3302,
  2022.

\bibitem{Tian2018AudioVisualEL}
Y.~Tian, J.~Shi, B.~Li, Z.~Duan, and C.~Xu, ``Audio-visual event localization
  in unconstrained videos,'' in \emph{European Conference on Computer Vision},
  2018.

\bibitem{8290832}
K.~R. Jerripothula, J.~Cai, and J.~Yuan, ``Efficient video object
  co-localization with co-saliency activated tracklets,'' \emph{IEEE
  Transactions on Circuits and Systems for Video Technology}, vol.~29, no.~3,
  pp. 744--755, 2019.

\bibitem{Lai2022MultiFeatureFB}
Z.~Q. Lai, Y.~Wang, R.~Feng, X.~Hu, and H.~Xu, ``Multi-feature fusion based
  deepfake face forgery video detection,'' \emph{Syst.}, vol.~10, p.~31, 2022.

\bibitem{Vaswani2017AttentionIA}
A.~Vaswani, N.~M. Shazeer, N.~Parmar, J.~Uszkoreit, L.~Jones, A.~N. Gomez,
  L.~Kaiser, and I.~Polosukhin, ``Attention is all you need,'' \emph{ArXiv},
  vol. abs/1706.03762, 2017.

\bibitem{46}
V.~Mnih, N.~Heess, A.~Graves, and K.~Kavukcuoglu, ``Recurrent models of visual
  attention,'' in \emph{NIPS}, 2014.

\bibitem{47}
J.~Ba, V.~Mnih, and K.~Kavukcuoglu, ``Multiple object recognition with visual
  attention,'' \emph{CoRR}, vol. abs/1412.7755, 2015.

\bibitem{48}
D.~Bahdanau, K.~Cho, and Y.~Bengio, ``Neural machine translation by jointly
  learning to align and translate,'' \emph{CoRR}, vol. abs/1409.0473, 2015.

\bibitem{49}
M.~Jaderberg, K.~Simonyan, A.~Zisserman, and K.~Kavukcuoglu, ``Spatial
  transformer networks,'' in \emph{NIPS}, 2015.

\bibitem{9439827}
C.~Wang, J.~Xue, K.~Lu, and Y.~Yan, ``Light attention embedding for facial
  expression recognition,'' \emph{IEEE Transactions on Circuits and Systems for
  Video Technology}, vol.~32, no.~4, pp. 1834--1847, 2022.

\bibitem{9393325}
H.~Wang, X.~Hu, X.~Zhao, and Y.~Zhang, ``Wide weighted attention multi-scale
  network for accurate mr image super-resolution,'' \emph{IEEE Transactions on
  Circuits and Systems for Video Technology}, vol.~32, no.~3, pp. 962--975,
  2022.

\bibitem{30}
S.~Woo, J.~Park, J.-Y. Lee, and I.-S. Kweon, ``Cbam: Convolutional block
  attention module,'' in \emph{ECCV}, 2018.

\bibitem{32}
N.~Srivastava, G.~E. Hinton, A.~Krizhevsky, I.~Sutskever, and R.~Salakhutdinov,
  ``Dropout: a simple way to prevent neural networks from overfitting,''
  \emph{J. Mach. Learn. Res.}, vol.~15, pp. 1929--1958, 2014.

\bibitem{Simonyan2014VeryDC}
K.~Simonyan and A.~Zisserman, ``Very deep convolutional networks for
  large-scale image recognition,'' \emph{CoRR}, vol. abs/1409.1556, 2014.

\bibitem{31}
S.~Ioffe and C.~Szegedy, ``Batch normalization: Accelerating deep network
  training by reducing internal covariate shift,'' \emph{ArXiv}, vol.
  abs/1502.03167, 2015.

\bibitem{Perez2017FiLMVR}
E.~Perez, F.~Strub, H.~de~Vries, V.~Dumoulin, and A.~C. Courville, ``Film:
  Visual reasoning with a general conditioning layer,'' in \emph{AAAI
  Conference on Artificial Intelligence}, 2017.

\bibitem{Han2006UnsupervisedEO}
J.~Han, K.~N. Ngan, M.~Li, and H.~Zhang, ``Unsupervised extraction of visual
  attention objects in color images,'' \emph{IEEE Transactions on Circuits and
  Systems for Video Technology}, vol.~16, pp. 141--145, 2006.

\bibitem{Lin2017VisualAttentionBasedBM}
Y.~Lin, Y.~Tong, Y.~Cao, Y.~Zhou, and S.~Wang, ``Visual-attention-based
  background modeling for detecting infrequently moving objects,'' \emph{IEEE
  Transactions on Circuits and Systems for Video Technology}, vol.~27, pp.
  1208--1221, 2017.

\bibitem{Liu2021DeepLB}
Y.~Liu, Z.~Zhang, X.~Liu, W.~Lei, and X.~Xia, ``Deep learning based mineral
  image classification combined with visual attention mechanism,'' \emph{IEEE
  Access}, vol.~9, pp. 98\,091--98\,109, 2021.

\bibitem{Li2018InIO}
Y.~Li, M.-C. Chang, and S.~Lyu, ``In ictu oculi: Exposing ai created fake
  videos by detecting eye blinking,'' \emph{2018 IEEE International Workshop on
  Information Forensics and Security (WIFS)}, pp. 1--7, 2018.

\bibitem{Sanderson2002TheVD}
C.~Sanderson, ``The vidtimit database,'' 2002.

\bibitem{34}
S.~Zhang, X.~Zhu, Z.~Lei, H.~Shi, X.~Wang, and S.~Li, ``S3fd: Single shot
  scale-invariant face detector,'' \emph{2017 IEEE International Conference on
  Computer Vision (ICCV)}, pp. 192--201, 2017.

\bibitem{35}
D.~P. Kingma and J.~Ba, ``Adam: A method for stochastic optimization,''
  \emph{CoRR}, vol. abs/1412.6980, 2015.

\bibitem{Szegedy2014GoingDW}
C.~Szegedy, W.~Liu, Y.~Jia, P.~Sermanet, S.~E. Reed, D.~Anguelov, D.~Erhan,
  V.~Vanhoucke, and A.~Rabinovich, ``Going deeper with convolutions,''
  \emph{2015 IEEE Conference on Computer Vision and Pattern Recognition
  (CVPR)}, pp. 1--9, 2014.

\bibitem{Selvaraju2016GradCAMVE}
R.~R. Selvaraju, A.~Das, R.~Vedantam, M.~Cogswell, D.~Parikh, and D.~Batra,
  ``Grad-cam: Visual explanations from deep networks via gradient-based
  localization,'' \emph{International Journal of Computer Vision}, vol. 128,
  pp. 336--359, 2016.

\end{thebibliography}
